\documentclass{article}

\usepackage[final]{corl_2021}

\usepackage{amsmath,amsfonts,bm}

\def\eqref#1{equation~\ref{#1}}

\def\1{\bm{1}}

\DeclareMathAlphabet{\mathsfit}{\encodingdefault}{\sfdefault}{m}{sl}
\SetMathAlphabet{\mathsfit}{bold}{\encodingdefault}{\sfdefault}{bx}{n}

\hypersetup{
  linkcolor  = blue,
  citecolor  = blue,
  urlcolor   = Aquamarine,
  colorlinks = true,
}
\usepackage{listings}
\usepackage{url}
\usepackage{soul}
\usepackage{booktabs}
\usepackage{multirow}
\usepackage{graphicx}
\usepackage{paralist}
\usepackage[font=scriptsize,labelfont=sl,labelsep=period]{caption}
\usepackage{subcaption}
\usepackage{wrapfig}
\usepackage{enumitem}
\usepackage[subtle]{savetrees}
\usepackage[dvipsnames]{xcolor}
\usepackage{pifont}  
\usepackage{comment} 
\usepackage{fmtcount} 
\usepackage[titletoc,title]{appendix}
\usepackage{authblk}
\usepackage{nicefrac}
\usepackage{bbm}
\usepackage{makecell}

\usepackage{titlesec}
\titlespacing*{\subsection}{0pt}{0.1\baselineskip}{0.05\baselineskip}
\titlespacing*{\section}{0pt}{0.2\baselineskip}{0.1\baselineskip}

\title{XIRL: Cross-embodiment Inverse Reinforcement Learning}

\makeatletter
\renewcommand\AB@affilsepx{, \protect\Affilfont}
\makeatother
\author[1,3\thanks{Work done as an intern at Google.}]{Kevin Zakka}
\author[2]{Andy Zeng}
\author[2]{Pete Florence}
\author[2]{Jonathan Tompson}
\author[1]{\\Jeannette Bohg}
\author[2]{Debidatta Dwibedi}
\affil[1]{Stanford University}
\affil[2]{Robotics at Google}
\affil[3]{UC Berkeley}

\newcommand{\ie}{i.e., }
\newcommand{\eg}{e.g., }

\definecolor{mygreen}{rgb}{0.032, 0.6392, 0.2039}
\usepackage{color}
\definecolor{deepblue}{rgb}{0,0,0.5}
\definecolor{deepred}{rgb}{0.6,0,0}
\definecolor{deepgreen}{rgb}{0,0.5,0}

\usepackage{listings}

\newcommand\pythonstyle{%
  \lstset{
    language=Python,
    basicstyle=\ttfamily,
    morekeywords={self},              
    keywordstyle=\bfseries\color{deepblue},
    emph={MyClass,__init__},          
    emphstyle=\ttb\color{deepred},    
    stringstyle=\color{deepgreen},
    frame=tb,                         
    showstringspaces=false,
  }%
}

\lstnewenvironment{python}[1][]
{
\pythonstyle
\lstset{#1}
}
{}

\begin{document}
\maketitle

\begin{abstract}
We investigate the visual \textit{cross-embodiment} imitation setting, in which agents learn policies from videos of other agents (such as humans) demonstrating the same task, but with stark differences in their \textit{embodiments} -- shape, actions, end-effector dynamics, etc.
In this work, we demonstrate that it is possible to automatically discover and learn vision-based reward functions from cross-embodiment demonstration videos that are robust to these differences.
Specifically, we present a self-supervised method for Cross-embodiment Inverse Reinforcement Learning (XIRL) that leverages temporal cycle-consistency constraints to learn deep visual embeddings that capture task progression from offline videos of demonstrations across multiple expert agents, each performing the same task differently due to embodiment differences.
Prior to our work, producing rewards from self-supervised embeddings typically required alignment with a reference trajectory, which may be difficult to acquire under stark embodiment differences.
We show empirically that if the embeddings are aware of task progress, simply taking the negative distance between the current state and goal state in the learned embedding space is useful as a reward for training policies with reinforcement learning. We find our learned reward function not only works for embodiments seen during training, but also generalizes to entirely new embodiments. Additionally, when transferring real-world human demonstrations to a simulated robot, we find that XIRL is more sample efficient than current best methods.
Qualitative results, code, and datasets are available at \href{https://x-irl.github.io}{\color{magenta}https://x-irl.github.io}
%
%
\end{abstract}


\section{Introduction}

The ability to learn new tasks from third-person demonstrations holds the potential to enable robots to leverage the vast quantities of tutorial videos that can be gleaned from the world-wide web (\eg YouTube videos). However, distilling diverse and unstructured videos into motor skills with vision-based policies can be daunting -- as the videos themselves are often not only captured from different camera viewpoints in different environments, but also with different experts that may use different tools, objects, or strategies to perform the same task. Perhaps most critically, there often exists a clear \textit{embodiment gap} between the human expert demonstrator, and the robot hardware that executes the learned policies. One approach to close this gap is to learn a mapping between the human and robot embodiment~\cite{argall2009survey}, which is a non-trivial intermediate problem in itself. 

Despite considerable progress in learning policies with paired observations and actions (\eg collected via teleoperation) \cite{zhang2018deep, florence2019self}, much less work has been done in getting robots to learn policies from tasks defined only by third-person observations of demonstrations \cite{sermanet2016unsupervised,sermanet2018time,schmeckpeper2020reinforcement}.
This is a surprisingly challenging problem, as different embodiments are likely to use unique strategies that suit them and allow them to make progress on a given task. For example, if asked to “place five pens into a cup”, a human hand is likely to scoop up all pens before slipping them into the cup, whereas a two-fingered gripper might instead need to pick and place each pen individually. Both strategies complete the task, but generate different state-action trajectories. In this setting, it may be difficult to acquire labeled frame-to-frame correspondences between expert demonstration videos and learned embodiments \cite{schmeckpeper2020reinforcement}, particularly across a multitude of embodiments or experts. The approach we investigate instead is whether we can successfully learn tasks through a learned notion of task progress that is invariant to the embodiment performing the task.

\begin{figure}[t]
    \centering
    \includegraphics[width=\textwidth]{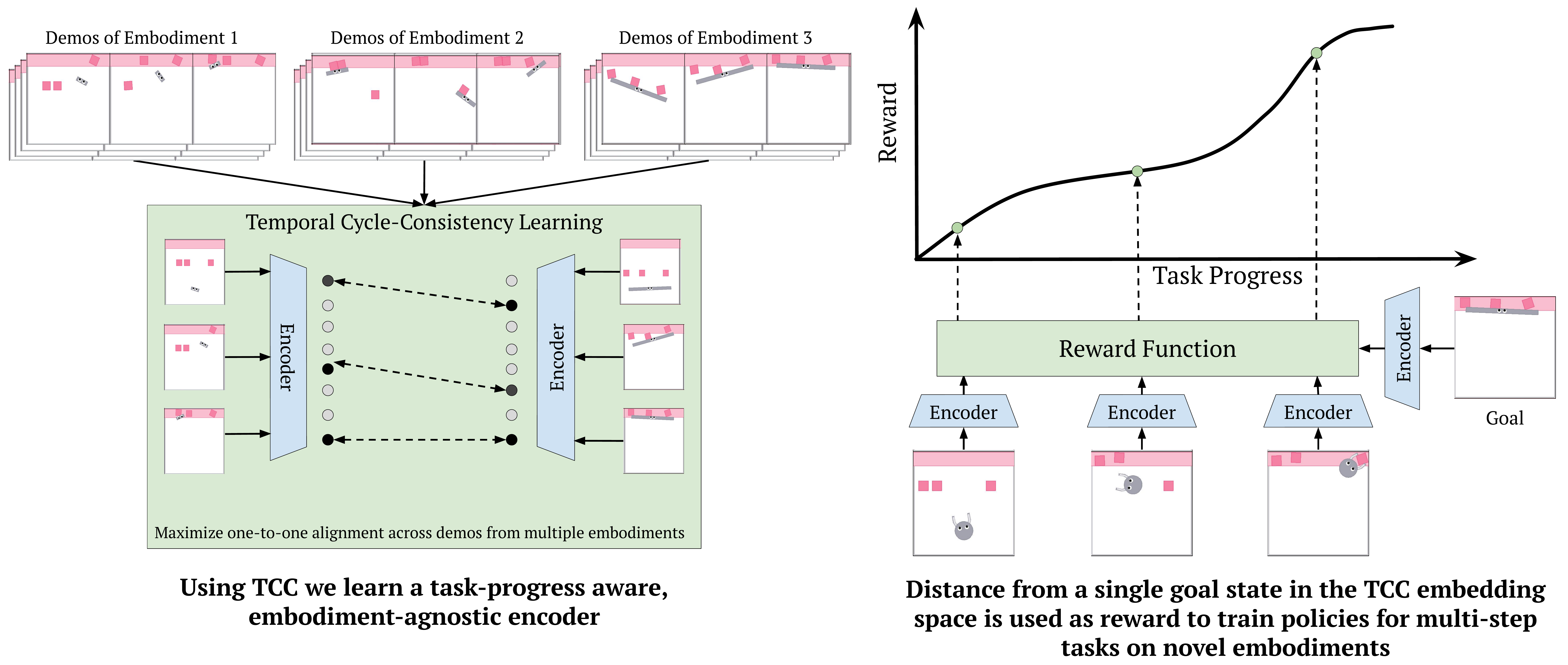}
    \caption{\textbf{Cross-embodiment Inverse Reinforcement Learning.} We learn embodiment-invariant visual representations from offline video demonstrations (stick agents on the left) using TCC \cite{Dwibedi_2019_CVPR}, then use the trained encoder to generate embodiment-invariant visual reward functions that can be used to learn policies on new embodiments (gripper on the right) with reinforcement learning.}
    \label{fig:embodiments}
    \vspace{-1em}
\end{figure}

In this work, we propose to enable agents to imitate video demonstrations of experts -- including ones with different embodiments -- by using a task-specific, embodiment-invariant reward formulation trained via temporal cycle-consistency (TCC) \cite{Dwibedi_2019_CVPR}. We demonstrate how we can leverage an encoder trained with TCC to define dense rewards for downstream reinforcement learning (RL) policies via simple distances in the learned embedding space.
Simulated experiments across four different embodiments show that these learned rewards are capable of generalizing to new embodiments, enabling unseen agents to learn the task via reinforcement, and surprisingly in some cases, exceeding the sample efficiency of the same agent learned with ground truth sparse rewards. We also demonstrate the effectiveness of our approach for learning robot policies using human demonstrations on the \textit{State Pusher} environment from \cite{schmeckpeper2020reinforcement}, where our reward is first learned on real-world human demonstrations, then used to teach a Sawyer arm how to perform the task in simulation.

Our contributions are as follows: \textbf{(i)} We introduce Cross-embodiment Inverse Reinforcement Learning (XIRL), an effective, label-free framework for tackling cross-embodiment visual imitation learning.  Our core contribution is to use self-supervised learning on third-person demonstration videos to define dense reward functions amenable for downstream reinforcement learning policies, \textbf{(ii)} Along with XIRL, we release a cross-embodiment imitation learning benchmark, X-MAGICAL, which features multiple simulated agents with different embodiments performing the same manipulation task, including one thousand expert demonstrations for each agent, \textbf{(iii)} We show that XIRL significantly outperforms alternative methods on both the X-MAGICAL benchmark and the human-to-robot transfer benchmark from \cite{schmeckpeper2020reinforcement}, and discuss our observations, which point to interesting areas for future research, \textbf{(iv)} Finally, we introduce a real-world dataset, X-REAL (Cross-embodiment Real demonstrations), of a manipulation task performed with nine different embodiments, which can be used to evaluate cross-embodiment reward learning methods\footnote{For more details regarding X-REAL, please see Appendix~\ref{appendix:x_real}.}.

\section{Related Work}

Traditional formulations of imitation learning \cite{Atkeson1997RobotLF,argall2009survey,osa2018algorithmicIL} assume access to a corpus of expert demonstration data which includes both state and action trajectories of the expert policy. In the context of third-person imitation learning, including when learning from expert agents with different embodiments, obtaining access to ground-truth actions is difficult or impossible\footnote{For a more comprehensive related work, please see Appendix~\ref{appendix:expanded_rw}.}.

\textbf{Inferring expert actions.} To address this issue, several approaches either try to infer expert actions \cite{torabi2018behavioral, edwards2018imitating, radosavovic2020state} -- for example by training an inverse dynamics model on agent interaction data~\cite{torabi2018behavioral} -- or employ forward prediction on the next state to imitate the expert without direct action supervision~\cite{pathak2018zero}.
While these methods successfully address learning from observation-only demonstrations, they either do not support skill transfer to different policy embodiments at all, or they cannot take advantage of multiple embodiments in order to improve generalization to unseen policy configurations. We explicitly address these problems in this work.


\textbf{Imitation via learned reward functions.} In contrast to imitation via supervised methods, such as BCO~\cite{torabi2018behavioral}, a recent body of work~\cite{sermanet2016unsupervised, sermanet2018time, aytar2018playing, schmeckpeper2020reinforcement, mees2020adversarial} has focused on learning reward functions from expert video data and then training RL policies to maximize this reward. In \cite{sermanet2016unsupervised}, the authors combine ImageNet pre-trained visual features with an L2-norm distance reward to match policy and expert observations in a latent feature space. In their follow-up work \cite{sermanet2018time}, the reward is computed in a viewpoint-invariant representation that is self-supervised on video data. While both these methods are compelling in their use of cheap unlabeled data to learn invariant rewards, the use of a time index as a heuristic for defining weak correspondence is a constraining limitation for tasks that need to be executed at different speeds, or are not strictly monotonic (\eg have ambiguous sub-task ordering).
In \cite{aytar2018playing} a dense reward is learned via unsupervised learning on YouTube data and the authors make no assumption about time alignment. However, in their work, the expert and learned policy are executed in the same domain and embodiment, an assumption we relax in our work. Framed in a multi-task learning setting, \cite{gupta2017learning} propose training policies with morphologically different embodiments first on a similar set of proxy tasks, in order to learn a latent space to map between domains, and then sharing skills on a held-out task from one policy to another. A time-index heuristic is used to define a metric reward when performing RL training of the new task. In our work, the learned embedding finds correspondences in a fully-unsupervised fashion, without the need for such strict time alignment. In \cite{singh2019end}, a small sub-set of states is human labeled for goal success and a convolutional network is then trained to detect successful task completion from image observations, where on-policy samples are used as negatives for the classifier. By contrast, our learned embedding encodes task progress in its latent representation without the use of expensive human labels.

\textbf{Imitation via domain adaptation.} An additional category of approaches to third-person imitation learning are those that perform domain adaptation of expert observations~\cite{smith2019avid, liu2018imitation, sharma2019third, hejna2020hierarchically}. For instance, in \cite{smith2019avid} a CycleGAN~\cite{zhu2017unpaired} architecture is used to perform pixel-level image-to-image translation between policy domains, which is then used to construct a reward function for a model-based RL algorithm. A similar model-free approach is proposed in \cite{liu2018imitation}. In \cite{sharma2019third}, a generative model is used to predict robot high-level sub-goals conditioned on a third-person demonstration video, and a lower-level control policy is trained in a task-agnostic manner. Similarly, \cite{hejna2020hierarchically} uses high level task conditioning from zero-shot transfer of expert demonstrations, but they use KL matching to perform both high and low-level imitation. In contrast to these methods, the unsupervised TCC alignment in this work avoids performing explicit domain adaptation or pixel-level image translation by instead learning a robust and invariant feature space in a fully offline fashion.


\textbf{Reinforcement learning with demonstrations.} Recent work in offline-reinforcement learning \cite{schmeckpeper2020reinforcement} explicitly tackles the problem of policy embodiment and domain shift. Their method, Reinforcement Learning from Videos (RLV), uses a labelled collection of expert-policy state pairs in conjunction with adversarial training to learn an inverse dynamics model jointly optimized with the policy. In contrast, we avoid the limitation of collecting human-labeled dense state correspondences by using a self-supervised algorithm (\ie TCC~\cite{Dwibedi_2019_CVPR}) which uses cycle-consistency to automatically learn the correspondence between states of two domains. We also show that this formulation improves generalization to unseen embodiments. Since the problem setup is similar to ours, we also compare to their method as a baseline.

\section{Approach}

Our overall XIRL framework (Figure~\ref{fig:embodiments}) addresses the \textit{cross-embodiment visual imitation} problem (Section~\ref{subsec:problem-formulation}). The framework consists of first using TCC to self-supervise embodiment-invariant visual representations (Section~\ref{subsec:representation-learning}), then using a novel embodiment-invariant visual reward function (Section~\ref{subsec:reward-learning}) to perform cross-embodiment visual imitation via  reinforcement learning  (Section~\ref{subsec:reinforcement-learning}).

\subsection{Problem Formulation}\label{subsec:problem-formulation}
Our objective is to extract an agent-invariant definition of a task, via a learned reward function, from a dataset of videos of separate agents performing the same task. In particular, we are interested in agents that may solve the task in entirely different ways due to differences in their end-effector, shapes, dynamics, etc., which we refer to as \textit{embodiment} differences. For example, consider how differently a vacuum gripper and a parallel-jaw gripper will grasp an object as a result of their respective end-effectors. Such a setup is quite common in robotic imitation learning, where we might have access to observation data of humans demonstrating a task, but want to teach a robot to perform it. It is very likely that the way the human executes the task will diverge from how the robot would execute it.

We define a dataset of multiple agents performing the same task $T$ as $D=\bigcup_{i=1}^n D_i$, where $D_i$ is an agent-specific dataset containing observations of \textit{only} agent $i$ performing task $T$. Each agent's dataset $D_i$ is a collection of videos defined as $D_i = \{v_i^1, v_i^2 ... v_i^K\}$, where $v_i^j$ represents the video of the $j^{th}$ demonstration of agent $i$ successfully performing the task $T$. We would like to highlight that $D$ only contains observation data, \ie it does not store the actions taken by the respective agents. We use self-supervised representation learning techniques to learn task-specific representations from this dataset.

\begin{figure*}[t!]
    \centering
    \begin{subfigure}[t]{0.3\textwidth}
        \centering
        \includegraphics[height=1.5in]{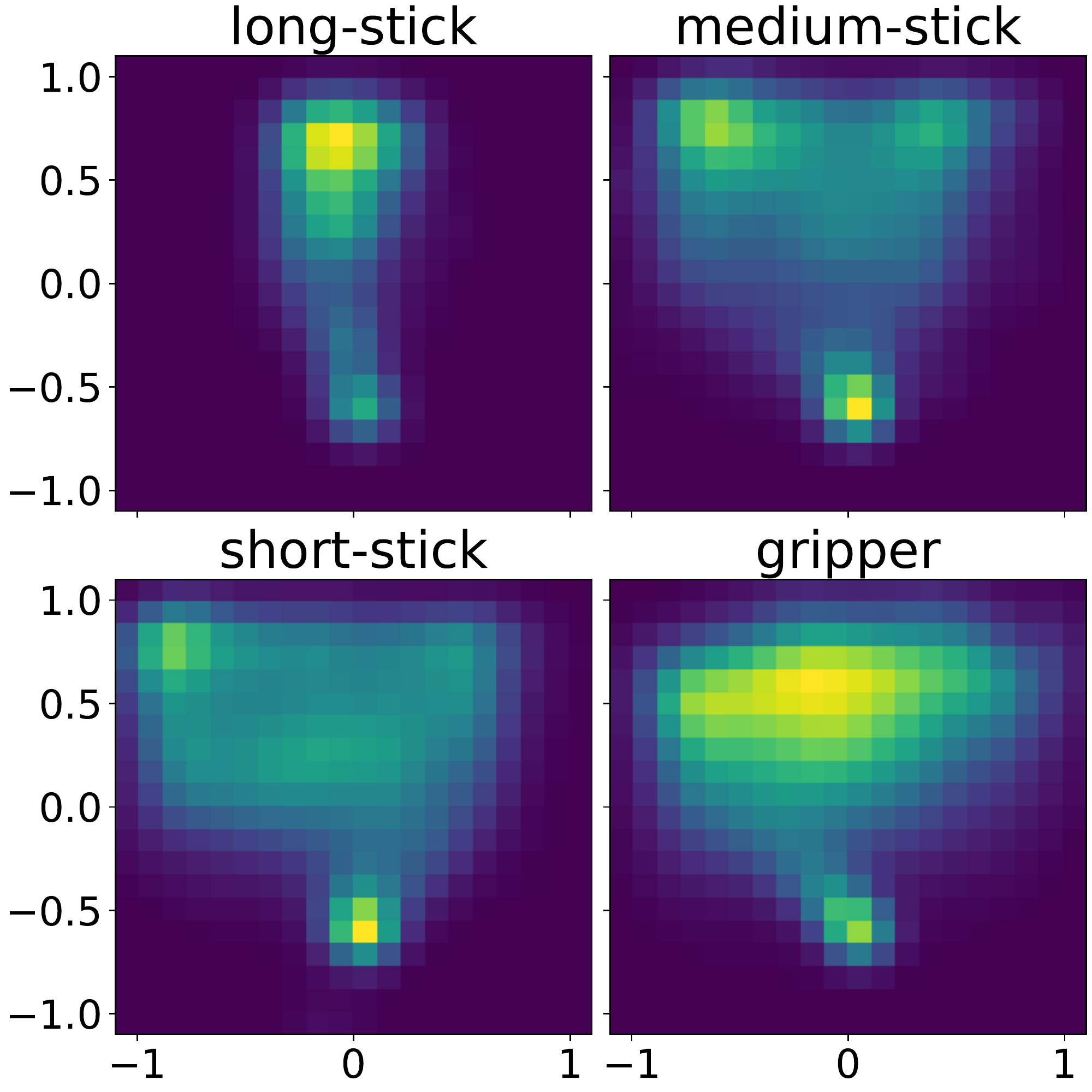}
        \caption{Agent state distribution.}
        \label{fig:different_agentsa}
    \end{subfigure}
    ~ 
    \begin{subfigure}[t]{0.3\textwidth}
        \centering
        \includegraphics[height=1.5in]{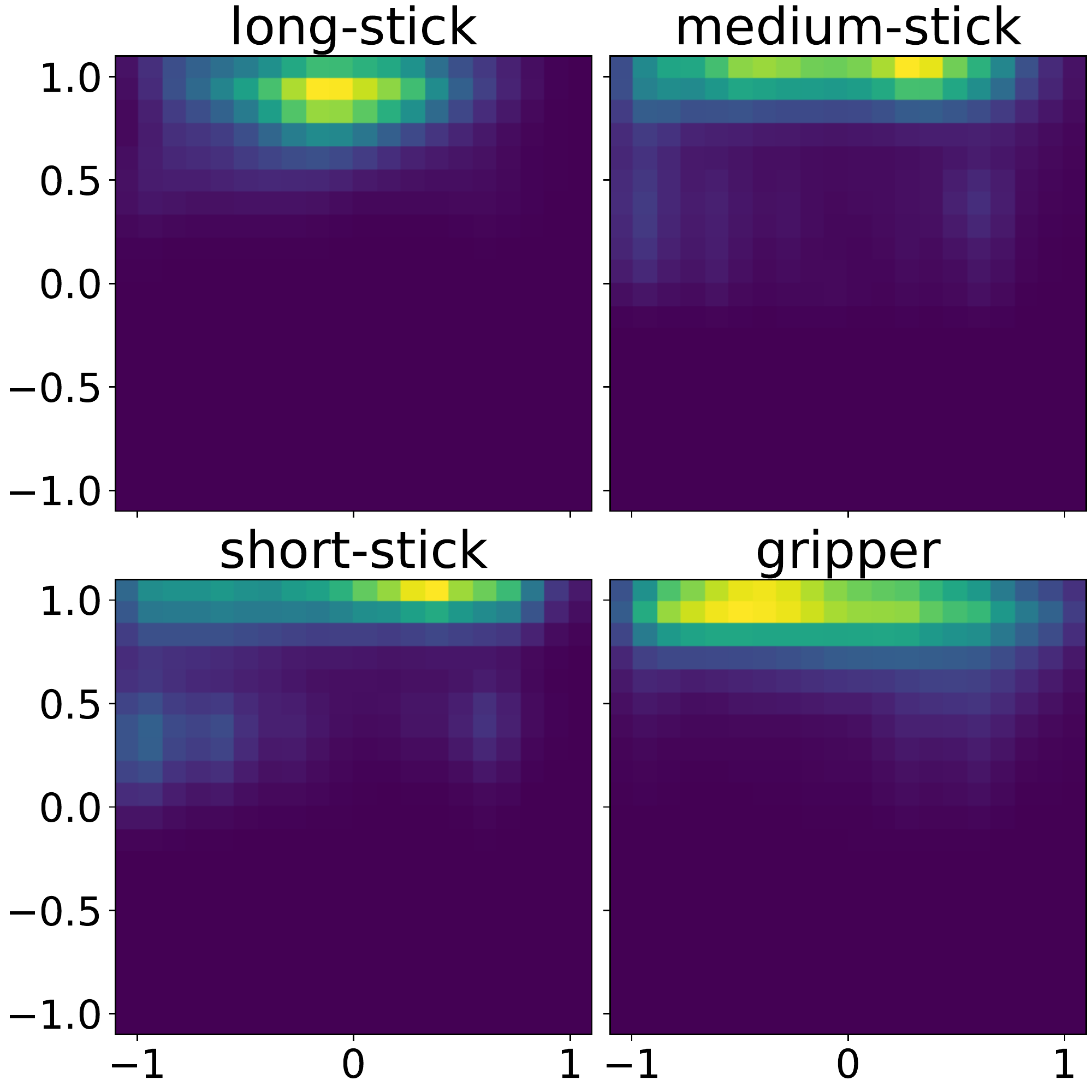}
        \caption{Debris state distribution.}
        \label{fig:different_agentsb}
    \end{subfigure}
    ~
    \begin{subfigure}[t]{0.3\textwidth}
        \centering
        \includegraphics[height=1.5in]{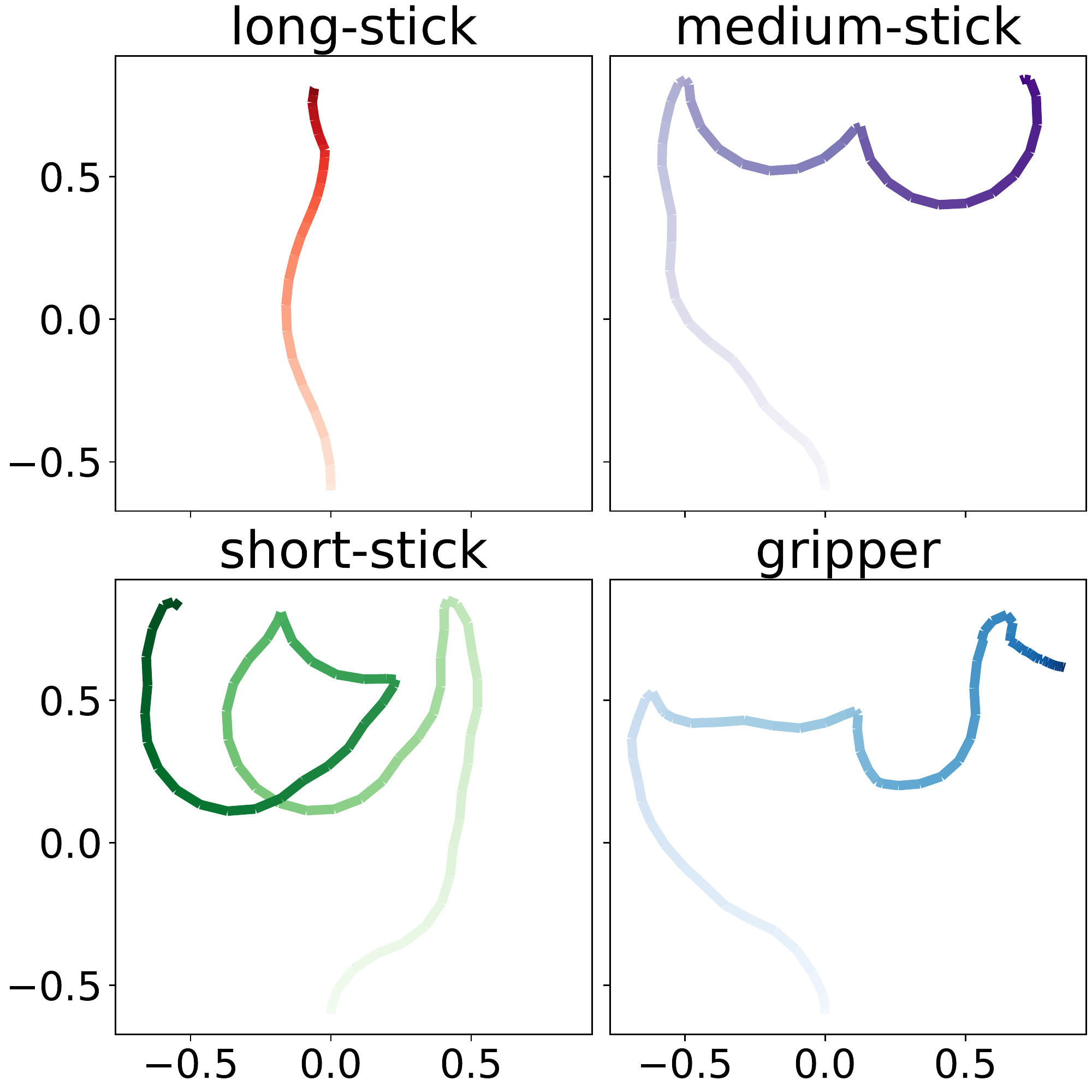}
        \caption{Agent trajectory in randomly sampled video.}
        \label{fig:different_agentsc}
    \end{subfigure}
    
    \caption{\textbf{Difference in state visitation distributions} across different embodiments performing the same task in X-MAGICAL (Section~\ref{subsection:x-magical}).}
    \label{fig:different_agents}
    \vspace{-1em}
\end{figure*}

\subsection{Representation Learning}\label{subsec:representation-learning}
In this work, we use TCC \cite{Dwibedi_2019_CVPR} to learn task-specific representations in a self-supervised way. The method has been shown to learn useful representations on videos of the same action for temporally fine-grained downstream tasks. In their paper, the authors show that TCC representations can predict frame-level task progress, such as predicting how much water is in a cup during pouring, without requiring any human annotations. Task progress can provide dense signals for learning a new task and we would like to bake this property into our learned reward. Another advantage of TCC is that it does not require supervision for frame-level alignment (\ie which frames in two videos correspond to each other). Such frame-to-frame correspondences are required by a prior method, RLV \cite{schmeckpeper2020reinforcement}, to achieve successful reinforcement learning on the considered tasks, but we would like to avoid this type of manual supervision.

We train an image encoder $\phi$, that ingests an image $I$ and returns an embedding vector $\phi(I)$, using TCC. TCC assumes that there exist semantic temporal correspondences between two video sequences, even though they are not labeled and the actions may be executed at different speeds. Note that in our case, the assumption holds since all sequences in our dataset originate from the same task. Furthermore, even if two different agents accomplish the task very differently, there will be a common set of states or frames that will temporally correspond. When we apply the TCC loss, we compare frames of one agent performing the task with frames from another, and search for temporal similarities in how both execute the task. By performing multiple such comparisons and \textit{cycling-back} (described in the next paragraph), TCC encodes task progress in its latent representation, a property that is useful for learning task-specific, yet embodiment-invariant reward functions. For completeness, we describe the training technique below.

We define a dataset $D_{\text{all}} = \{v_1, v_2 ... v_N\}$ that contains the videos of all agents executing the same task $T$. We are able to merge the datasets of different agents since TCC does not require embodiment labels (\ie IDs) during training. We first sample a random batch of videos and embed all their frames using the aforementioned image encoder $\phi$. For each video $v_i$, this results in a sequence of embeddings $V_i = \{\phi(v_i^1), \phi(v_i^2) ..., \phi(v_i^{L_i}) \}$, where $L_i$ is the length of the $i^{th}$ video. From this mini-batch of sequences of video frame embeddings, we choose a pair of sequences $V_i$ and $V_j$ and compute their TCC loss. In particular, we randomly sample a frame embedding from sequence $V_i$ -- say $V_i^t$ corresponding to the $t^{th}$ frame of video $v_i$ -- and compute the soft nearest-neighbor of $V_i^t$ in sequence $V_j$ in the embedding space as follows: 
\begin{equation*}
\widetilde{V_{ij}^t} = \sum_k^{L_j} \alpha_k V_j^k, \quad \mathrm{where} \quad \alpha_k = \frac{e^{-\Vert V_i^t-V_j^k\Vert^2}}{\sum_k^{L_j} e^{-\Vert V_i^t-V_j^k \Vert^2}}
\end{equation*} 
We then \textit{cycle-back} \cite{Dwibedi_2019_CVPR} to the first sequence $V_i$ by computing the soft-nearest neighbor of $\widetilde{V_{ij}^t}$ with all the frames in $V_i$. The probability of cycling-back to the $k^{th}$ frame in $V_i$ can be computed as:
\begin{equation*}
\beta_{ijt}^{k} = \frac{e^{-\Vert\widetilde{V_{ij}^t}-V_i^k\Vert^2}}{\sum_k^{L_i} e^{-\Vert\widetilde{V_{ij}^t} - V_i^k\Vert^2}}
\end{equation*}
The expected frame index we cycle-back to is then $\mu_{ij}^{t} =\sum_k^{L_i} \beta_{ijt}^k k$.
Since we know the index of the frame that started the cycle, in this case $t$, we can minimize the mean-squared (MSE) error loss between $t$ and the closest index retrieved via soft-nearest neighbor, \ie $\widetilde{V_{ij}^t}$. The loss for a single frame is thus: $L_{ij}^{t} = (\mu_{ij}^{t} - t)^2$.
Finally, we minimize the average loss $L$ over all frames in video $v_i$ with all other videos in the dataset $v_j$, defined as $L = \sum_{ijt} L_{ij}^t$.

    

\subsection{Reward Function}\label{subsec:reward-learning}
Once the encoder $\phi$ has been trained on demonstrations of different agents performing the same task $T$, we want to use it to transfer information about the task from one agent to another. We do this by leveraging $\phi$ to generate rewards via distances to goal observations in the learned embedding space.
Specifically, we define the goal embedding $g$ as the mean embedding of the last frame of all the demonstration videos in our offline dataset $D_{\text{all}}$. Concretely, $g = \nicefrac{\sum_{i=1}^N \phi(v_i^{L_i})}{N}$, where $L_i$ is the length of video $v_i$. Our reward $r$ then is the scaled negative distance of the current state embedding to the goal embedding $g$ \ie $r(s) = \nicefrac{-1}{\kappa} \cdot \Vert \phi(s) - g \Vert^2_2 ,$
%
%
where $\phi(s)$ is the state embedding at the current timestep and $\kappa$ is a scale parameter that ensures the distances are in a range amenable for training reinforcement learning algorithms~\cite{henderson2018deep}. We found it effective to set $\kappa$ to be the average distance of the first frame's embedding to the goal embedding for all the demonstrations in the dataset.
Defining $r$ in this manner gives us several advantages: (a) it is dense, encoding both task completion and task progress, (b) it does not require any correspondence with a reference trajectory~\cite{sermanet2018time, aytar2018playing} and (c) it sidesteps the need for a finite library of reference trajectories, unlike prior work \cite{sermanet2018time, gupta2017learning} that define time-indexed rewards relative to some reference trajectory. Thus, agents with trajectories of varying lengths (due to embodiment-specific constraints) can efficiently leverage this reward because the learned encoder can map different strategies to a common notion of task progress.

\subsection{Reinforcement Learning}\label{subsec:reinforcement-learning}
Using the pre-trained frozen encoder $\phi$, we define a Markov Decision Process (MDP) for any agent as the tuple $\langle \mathcal{S},~\mathcal{A},~P,~r \rangle$ where $\mathcal{S}$ is the set of possible states, $\mathcal{A}$ is the set of possible actions, $P$ is the state transition probability matrix encoding the dynamics of the environment (including the agent) and $r$ is the \textit{learned} reward function (defined in Section~\ref{subsec:reward-learning}). Notice how we are able to use the same reward function $r$ for any agent -- even ones that the encoder may not have seen during training. Furthermore, this reward function solely depends on the learned encoder. Hence, the task represented by the MDP now depends solely on how well the encoder has learned task-specific representations since we do not use the environment reward to either define the task or learn the policy. This is an important distinction because we are expecting the encoder to generalize to new states that an agent might encounter during training, as the expert demonstrations in our dataset only contain successful trajectories. It is also possible to augment sparse rewards (which only define task success or failure) with our learned dense reward while training policies.

\begin{table}[t]
\centering
\caption{Statistics of Demos of the X-MAGICAL Embodiments}
\footnotesize
\begin{tabular}{l|c}
\toprule
\textbf{Embodiment}          & \textbf{Mean $\pm$ Std. Dev. Demo Length} \\
     & \textbf{(no. of frames)} \\
\midrule
\midrule
Long-stick   & $\phantom{0}48.0\phantom{0}  \pm 28.5\phantom{0}$ \\
Medium-stick & $\phantom{0}59.9\phantom{0}  \pm 27.9\phantom{0}$ \\
Short-stick  & $\phantom{0}81.7\phantom{0}  \pm 35.0\phantom{0}$ \\
Gripper      & $110.8\phantom{0} \pm 39.8\phantom{0}$ \\
\bottomrule
\end{tabular}
\label{tab:stats}
\vspace{-1em}
\end{table}

\section{Experimental Setup}

\subsection{X-MAGICAL Benchmark}\label{subsection:x-magical}
We introduce a cross-embodiment imitation learning benchmark, X-MAGICAL, which is based on the imitation learning benchmark MAGICAL~\cite{toyer2020magical}, implemented on top of the physics engine PyMunk~\cite{pymunk}. In this work, we consider a simplified 2D equivalent of a common household robotic sweeping task, wherein an agent has to push three objects into a predefined zone in the environment (colored in pink). We choose this task specifically because its long-horizon nature highlights how different agent embodiments can generate entirely different trajectories. The reward in this environment is defined as the fraction of debris swept into the zone at the end of the episode.

\textbf{Multiple Embodiments in X-MAGICAL.}
We create multiple \textit{embodiments} by designing agents with different shapes and end-effectors that induce variations in how each agent solves a task. In Figure~\ref{fig:embodiments}, we show three of these embodiments and some sample trajectories that solve the sweeping task. Please see Appendix~\ref{appendix:x_magical} for a detailed description of the benchmark. Three agents are shaped like a \textit{stick} and they differ only in length. We call them \textit{short-stick}, \textit{medium-stick} and \textit{long-stick} based on the length of their body. The agent in the last row is called \textit{gripper}: it is circular in shape and has two arms that can actuate. All agents are capable of two actions - a rotation around their axis and a translation in a forward/backward direction along this axis (similar to the agent in the MAGICAL benchmark). All agents have a two-dimensional action space and use force and torque control to change their position and orientation respectively. The gripper agent has an additional degree of freedom for opening or closing its fingers. The default state of the gripper's fingers is open. For all agents, the state representation is a 16-dimensional vector with the following information: $(x,~y)$ position of the agent, $(\cos~{\theta},~\sin~{\theta})$ where $\theta$ is the agent's 2D orientation, and for each of the three debris: its $(x,~y)$ position, its distance to the agent and its distance to the goal zone. We frame-stack~\cite{mnih2015human} three consecutive state vectors to encode temporal and velocity information, resulting in a final state dimension of $48$.
 

\textbf{Demonstrations and Different Embodiment Strategies in X-MAGICAL.}
To learn task-specific representations for this task, we collect 1000 demonstrations per agent, where each demonstration consists in sweeping all three debris, initialized with random positions, into the target zone. This is the dataset $D_{\text{all}}$ (described in Section~\ref{subsec:problem-formulation}) containing observation-only agent-specific demonstrations. In Figure~\ref{fig:different_agents}, we highlight the differences that exist between the trajectories taken by these agents. Figure~\ref{fig:different_agentsa} shows a heat map of the frequency of visits (\textit{state visitation count}) of each agent at every 2D position in the grid, across all demonstrations in the dataset. We plot the 2D projection of the state visitation count onto the $XY$ plane with a bin width of $0.1$. Yellow encodes higher state visitation whereas blue encodes lower state visitation. We observe that agent \textit{long-stick} has less coverage of the environment as opposed to agent \textit{gripper}, which has significantly more coverage. Similarly, we show the distribution of debris locations for all agents across all demonstrations in Figure~\ref{fig:different_agentsb}. In Figure~\ref{fig:different_agentsc}, we plot a randomly sampled trajectory from each agent's demonstration pool. We use transparency to encode the start (lighter) and end (darker) of the trajectory. It is clear from this figure that each agent solves the task in a different manner. Additionally, there is a significant difference in the time taken by each agent to execute this task, as shown in Table~\ref{tab:stats}. Agent \textit{long-stick} is able to finish the task the quickest because of its long shape that can sweep all the debris at once, while agents \textit{short-stick} and \textit{gripper} take longer because they have to frequently push or grasp one debris at a time. These differences are the types of challenges that a representation must overcome to successfully generalize across embodiments. As such, X-MAGICAL serves to create a highly simplified version of a real-world scenario where we might want to learn new tasks from an observation dataset of humans performing these tasks in highly diverse ways.

\subsection{Baselines}
\label{sec:baselines}
\begin{figure}[t]
    \centering
    \includegraphics[width=\textwidth]{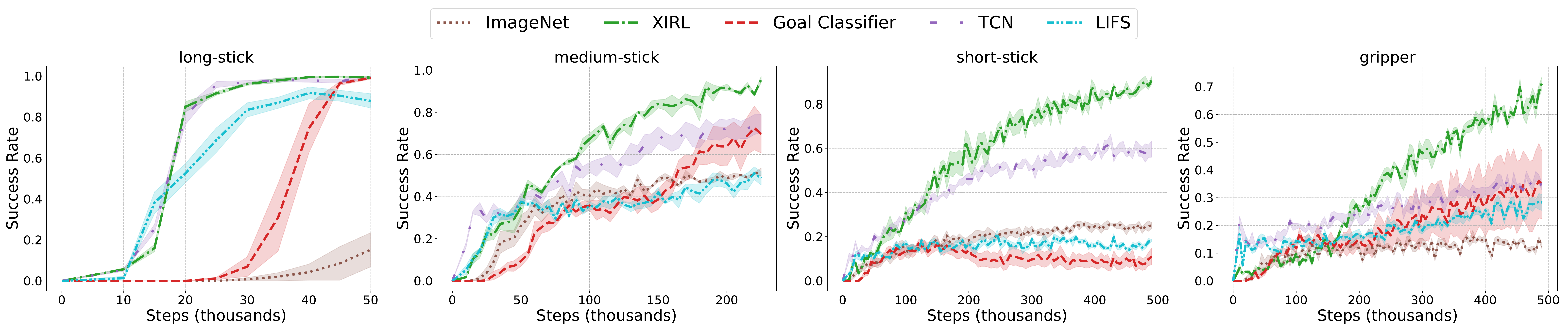}
    \caption{\textbf{\textit{Same-embodiment setting:}} Comparison of XIRL with other baseline reward functions, using SAC \cite{haarnoja2018soft} for RL policy learning on the X-MAGICAL \textit{sweeping} task.}
    \label{fig:exp1}
    \vspace{-2em}
\end{figure}

Here, we describe the alternative reward functions we baseline our method against, color coded to match their appearance in Figs. \ref{fig:exp1}, \ref{fig:exp3a}, and \ref{fig:rlv_push}.
%
%
\noindent\textcolor{brown}{\textit{1) ImageNet}:} We use an ImageNet pre-trained ResNet-18 encoder with no additional self-supervised training, \ie we load the pre-trained weights, discard the classification head, and use the 512-dimensional embedding space from the previous layer.
\noindent\textcolor{red}{\textit{2) Goal classifier}:} We follow \cite{vecerik2019practical} and train a goal frame classifier on a binary classification task where the last frame of all the demonstrations is considered positive and all the others are considered negatives.
We use the output probabilities of the classifier as the reward function.
\noindent\textcolor{cyan}{\textit{3) LIFS}:} We implement the method from \cite{gupta2017learning} which learns a feature space that is invariant to different embodiments using a contrastive loss function paired with an autoencoding loss.
%
\noindent\textcolor{violet}{\textit{4) TCN}:} single-view Time-Contrastive Network (TCN) \cite{sermanet2018time} with positive and negative frame windows of $1$ and $4$ respectively.
For more details regarding baseline implementations, see Appendix~\ref{app:baseline_details}.

\subsection{Implementation Details}
Our encoder for all experiments and methods is a ResNet-18 \cite{He2016resnet} initialized with ImageNet pre-trained weights. We replace the classification head with an embedding layer outputting a 32-dimensional vector. The encoder is trained on images of resolution $224 \times 224$ with ADAM \cite{kingma2014adam} and a learning rate of $10^{-5}$.
Note that our learned reward is agnostic to the RL algorithm used -- in this work, we opt for Soft-Actor Critic (SAC) \cite{haarnoja2018soft}, which is a  reinforcement learning algorithm that has been successfully used to train policies for continuous control tasks \cite{haarnoja2018sacapps}. Once the TCC encoder is trained, we use it to embed the observation frames as the agent interacts with the environment.

\section{Experiments}
We execute a series of experiments to evaluate whether the learned reward functions are effective at visual imitation. Specifically, our experiments seek to answer the following questions: first (Section~\ref{subsection:same-embodiment}), in the \textbf{\textit{same}}-embodiment case, where the demonstration dataset $D$ contains the embodiment of the learning agent, does our method enable successful reinforcement learning for that agent?
Next (Section~\ref{subsection:cross-embodiment}), we
investigate our primary interest, the \textbf{\textit{cross}}-embodiment case, where the demonstration dataset $D$ does \textbf{not} contain the embodiment of the learning agent. To additionally test our approach using real-world data (Section~\ref{subsection:real-cross-embodiment}), we use the dataset from \cite{schmeckpeper2020reinforcement} to leverage real-world human demonstrations to learn policies in simulation.
Note that each embodiment's performance is evaluated over 50 episodes and all figures plot the mean performance over 5 random seeds, with a standard deviation shading of $\pm 0.5$. Videos of our results are in the supplementary video.


\begin{figure}
    \centering
    \includegraphics[width=\textwidth]{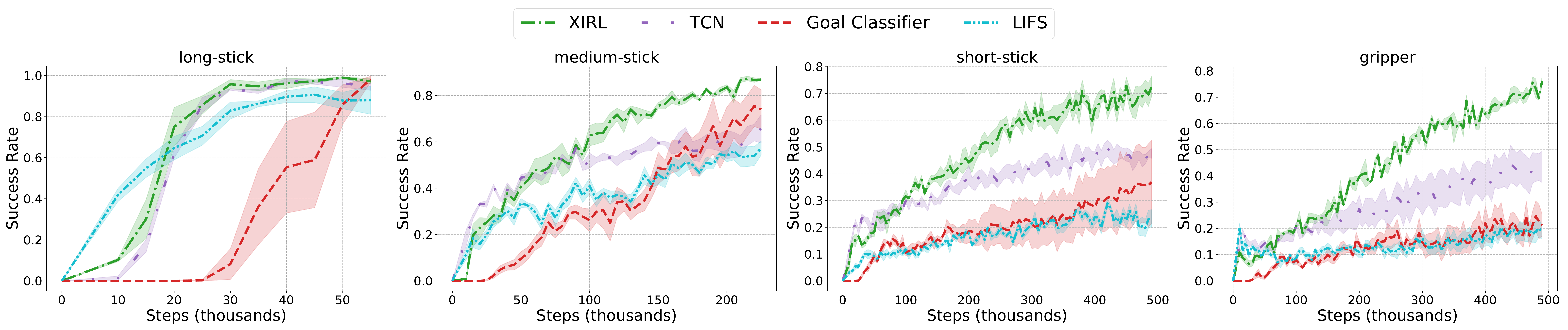}
    \caption{\textbf{\textit{Cross-embodiment setting:}} XIRL performs favorably when compared with other baseline reward functions, trained on observation-only demonstrations from different embodiments. Each agent \{\textit{long-stick, medium-stick, short-stick, gripper}\} is shown using demonstrations from the other 3 embodiments, with SAC \cite{haarnoja2018soft} for RL policy learning on the X-MAGICAL \textit{sweeping} task. 
    }
    \label{fig:exp3a}
    \vspace{-2em}
\end{figure}

\subsection{Results on Learning from \textbf{Same}-Embodiment Demonstrations}\label{subsection:same-embodiment}
In this experiment, we want to validate whether our approach of using a  reward function trained with TCC is good enough to train agents to perform the task defined in a dataset of expert demonstrations. Note that the learned reward function has to be robust enough such that it can provide a useful signal for new states an agent might encounter while learning a policy and interacting with the environment.
%
%
In Figure~\ref{fig:exp1}, we compare our method XIRL with baselines described in Section~\ref{sec:baselines}. We find XIRL is more sample-efficient than the other learned reward baselines. We attribute this sample efficiency to the fact that the TCC embeddings encode task progress which helps the agent learn to reach for objects and goal zones while interacting with the environment, rather than exploring in a purely random manner. This experiment provides evidence that XIRL's reward function is suitable for downstream reinforcement learning.

\subsection{Results on Learning from \textbf{Cross}-Embodiment Demonstrations}\label{subsection:cross-embodiment}
After verifying that XIRL works on the embodiments it was trained with, we move to the experiments that answer the \textbf{core question} in our work: \textit{can XIRL generalize to unseen embodiments}? In this section we conduct experiments where the reward function is learned using embodiments that are different from the ones on which the policy is trained. As noted in Table~\ref{tab:stats}, the timescales with which the agents execute the task can vary significantly.
We conduct four experiments, each one corresponding to holding-out one agent from the expert demonstration set. In each such experiment, we train an encoder on demonstrations from the remaining three agents. We compare with reward functions learned using TCN, LIFS, and goal frame classifiers. While both TCC and TCN are contrastive losses, the former makes explicit comparisons across different embodiments, whereas the latter implicitly relies on an encoder shared across embodiments to learn the cross-embodiment representation. In Figure~\ref{fig:exp3a}, we show that XIRL generalizes to new agents significantly better than the other learned reward baselines.


\subsection{Results on Learning from \textbf{Real-World Cross}-Embodiment Demonstrations}
\label{subsection:real-cross-embodiment}

\begin{figure}
    \centering
    \includegraphics[width=.38\textwidth,height=1.5in]{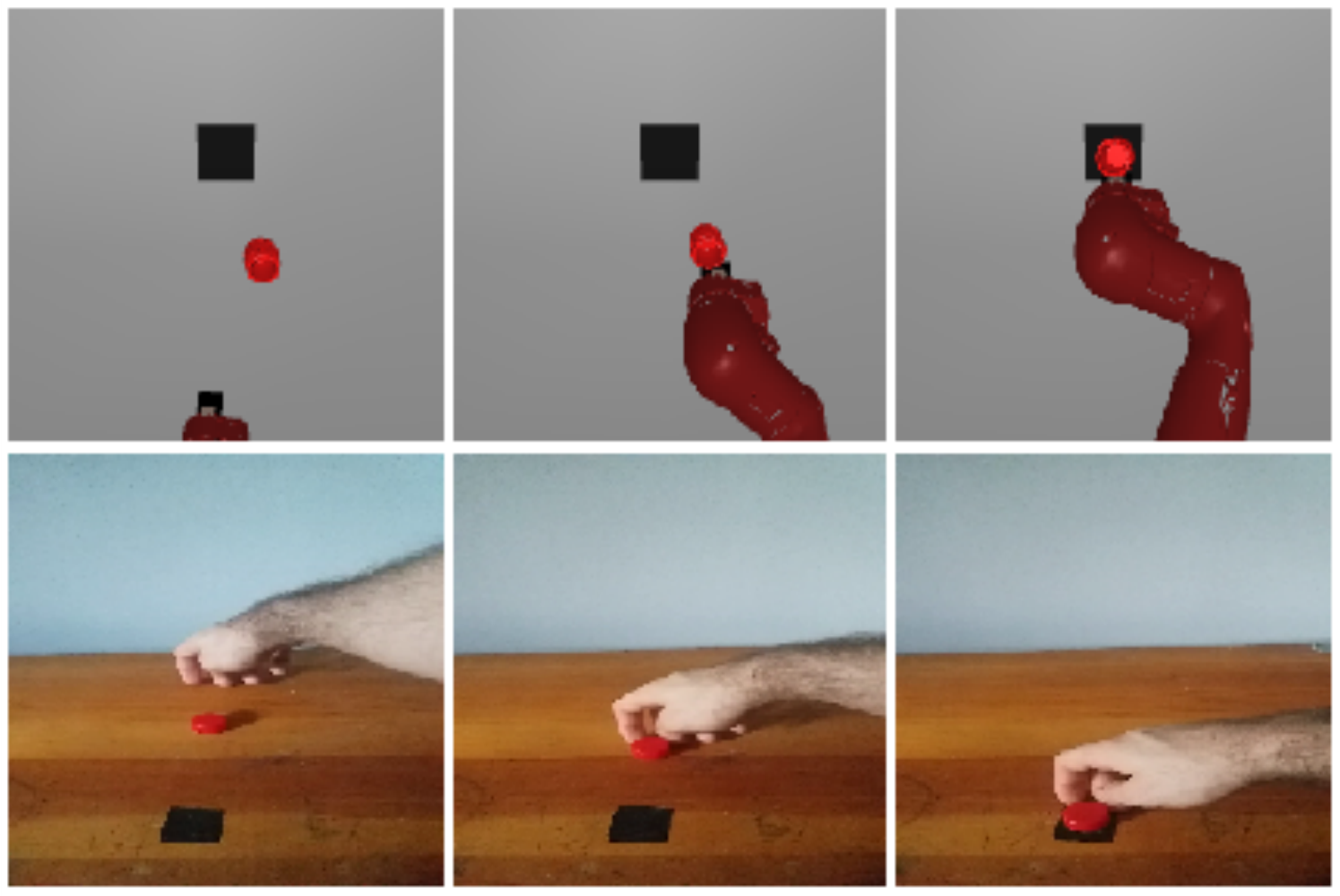}
    \includegraphics[width=.38\textwidth,height=1.5in]{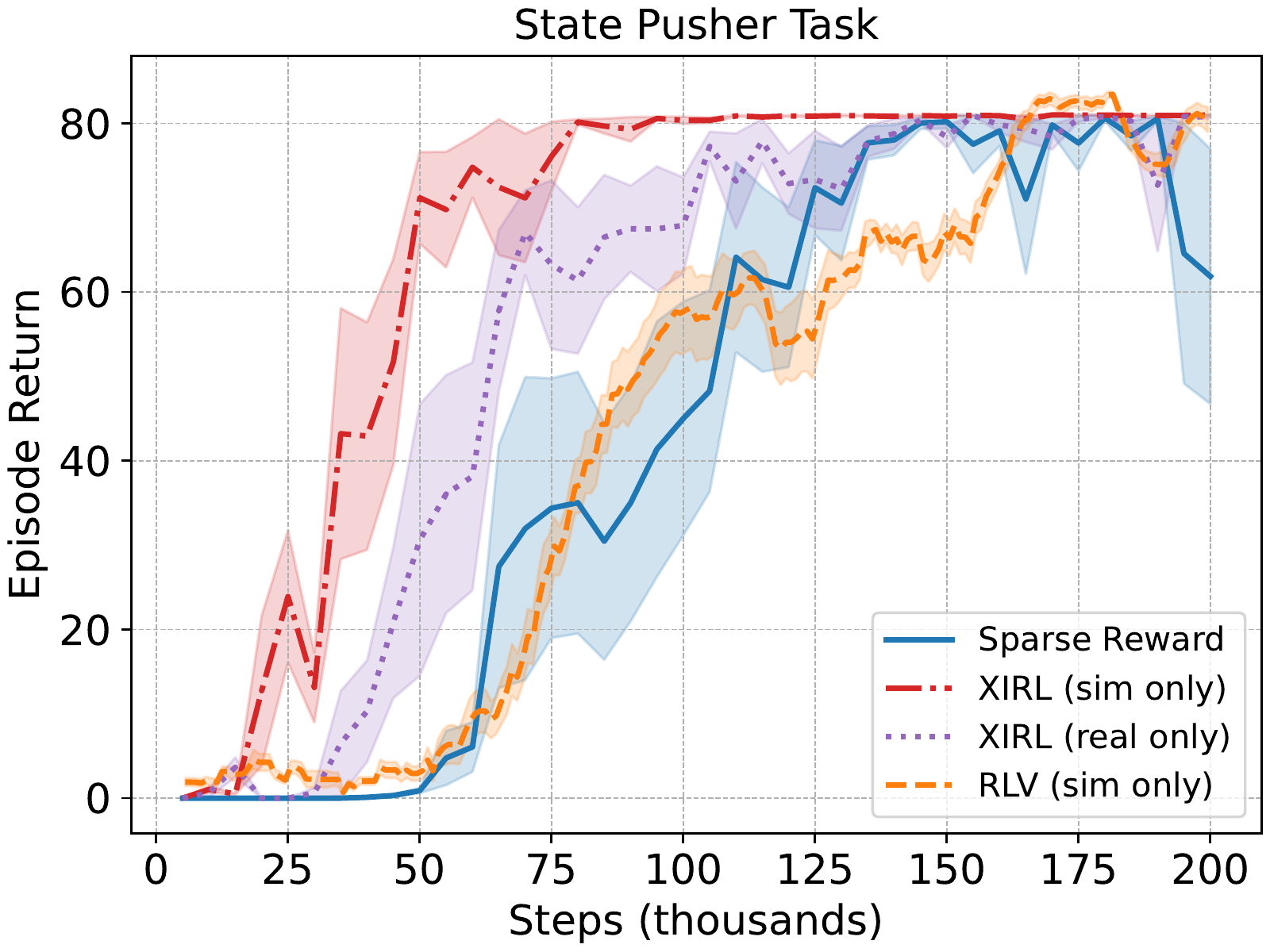}
    \caption{\textbf{\textit{Real-world-demo cross-embodiment setting:}} Comparison of XIRL with baselines using the simulated \textit{State Pusher} environment from \cite{schmeckpeper2020reinforcement}. XIRL (real only) leverages real-world demonstration videos of humans (left, row 2) to teach a robot arm in sim (left, row 1), but unlike \cite{schmeckpeper2020reinforcement}, we do not use human-labeled data of paired frame correspondences.
    RLV* denotes results taken verbatim from \cite{schmeckpeper2020reinforcement} which uses a different implementation of SAC for RL policy learning.}
    \label{fig:rlv_push}
\end{figure}

In this experiment, we test how well we can learn rewards from more challenging real-world human demonstration videos. To do so, we use the dataset and \textit{State Pusher} environment introduced in \cite{schmeckpeper2020reinforcement}. We train two XIRL encoders: XIRL (sim only) trained on 5 teleoperated simulated trajectories (\ie no domain shift) and XIRL (real only) trained solely on the real-world human demonstrations without using any form of human labeling of paired frames. We compare our results to training the policy on the sparse reward from the environment and the RLV method presented in \cite{schmeckpeper2020reinforcement}. As demonstrated in Figure \ref{fig:rlv_push}, we find our approach can improve the sample-efficiency of learning in the environment, compared to using RLV or solely the environment reward.

\subsection{Qualitative comparison between learned reward functions vs. handcrafted rewards}
\begin{figure}[ht]
    \centering
    \includegraphics[width=0.6\textwidth]{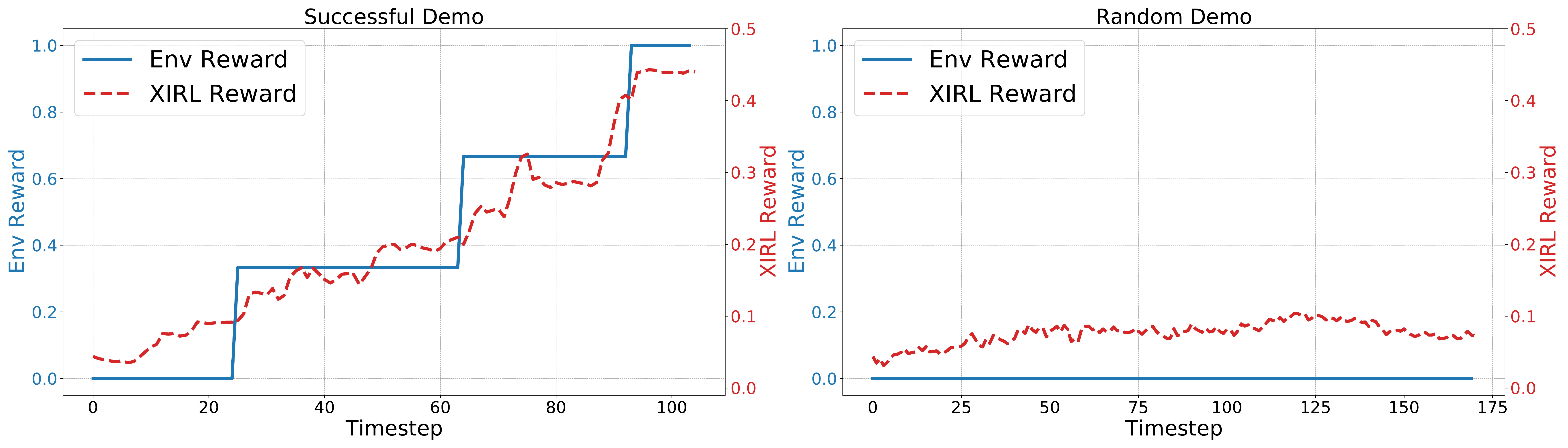}
    \caption{\textbf{X-MAGICAL demo cross-embodiment setting:} Visualizing our learned reward function XIRL vs. the environment's sparse reward on a successful and unsuccessful demonstration, for the \textit{short-stick} agent on the X-MAGICAL \textit{sweeping} task. The learned reward was trained on the three other agents.}
    \label{fig:qual_reward_plots_magical}
    \vspace{-1em}
\end{figure}

\begin{figure}[ht]
    \centering
    \includegraphics[width=0.8\textwidth]{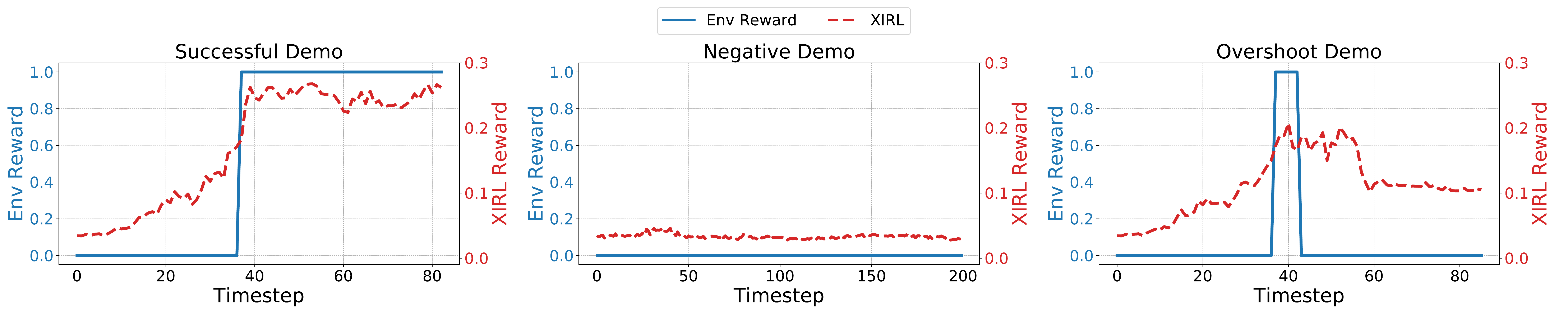}
    \includegraphics[width=0.8\textwidth]{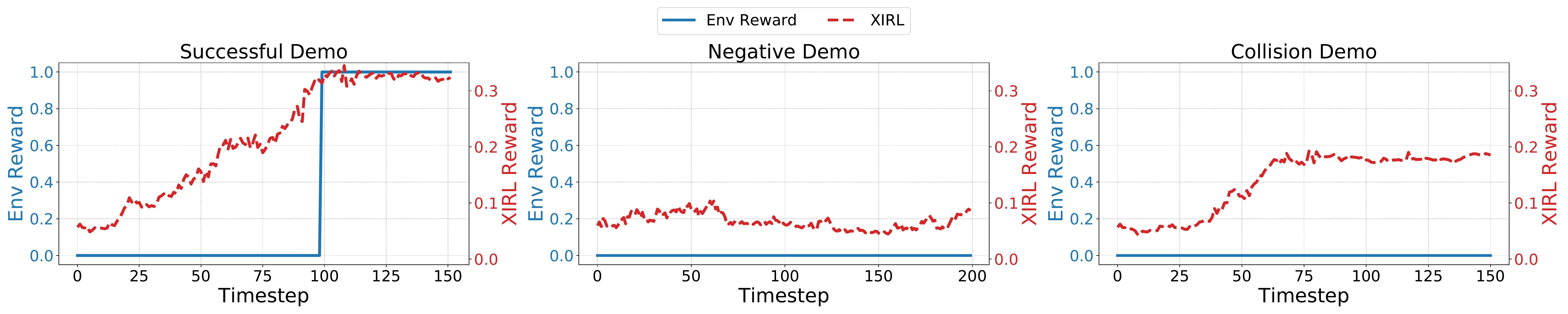}
    \caption{\textbf{\textit{Real-world-demo cross-embodiment setting:}} Visualizing our learned XIRL reward function vs. the environment's sparse reward on the \textit{State Pusher} (top) and \textit{Drawer Opening} (bottom) tasks from \cite{schmeckpeper2020reinforcement}.}
    \label{fig:qual_reward_plots_drawer}
    \vspace{-1em}
\end{figure}

In Figures~\ref{fig:qual_reward_plots_magical} and~\ref{fig:qual_reward_plots_drawer}, we visualize and compare the XIRL reward function with the environment's reward (ground truth) for the \textit{Sweeping} task from X-MAGICAL (see Sec.~\ref{subsection:x-magical} for details) and the \textit{State Pusher} and \textit{Drawer Opening} tasks from \cite{schmeckpeper2020reinforcement}\footnote{For additional qualitative visualizations on a real-world dataset with multiple embodiments, see Appendix~\ref{appendix:x_real}.}. We find that the learned reward is highly correlated  with the ground truth reward from the environment for both successful demos (first column in Figures~\ref{fig:qual_reward_plots_magical} and~\ref{fig:qual_reward_plots_drawer}) and unsuccessful demos (second column in Figures~\ref{fig:qual_reward_plots_magical} and~\ref{fig:qual_reward_plots_drawer}). It is especially encouraging to see that in a sparser reward environment like the \textit{Drawer Opening} task, XIRL provides a dense signal that should allow the agent to learn the task more efficiently.  Additionally, we can see that in the example of the failed collision trajectory (\ie second row third column in Fig. ~\ref{fig:qual_reward_plots_drawer}), where the arm collides with the drawer rather than opening it, XIRL is able to provide it with a partial reward (\ie for correctly moving towards the drawer) as opposed to the environment reward which remains zero.

\label{sec:xreal}
  
\section{Conclusion} 
\label{sec:conclusion}

This paper presents XIRL, a framework for learning vision-based reward functions from videos of expert demonstrators exhibiting different embodiments. XIRL uses TCC to self-supervise a deep visual encoder from videos, and uses this encoder to generate rewards via simple distances to goal observations in the embedding space. XIRL enables unseen agents with new embodiments to learn the demonstrated tasks via IRL. Reward functions from XIRL are fully self-supervised from videos, and we can successfully learn tasks without requiring manually paired video frames \cite{schmeckpeper2020reinforcement} between the demonstrator and learner.
In this sense, our method presents favorable \textit{scalability} to an arbitrary number of embodiments or experts with varying skill levels.
Experiments show that policies learned via XIRL are more sample efficient than multiple baseline alternatives, including TCN~\cite{sermanet2018time}, LIFS~\cite{gupta2017learning}, and RLV~\cite{schmeckpeper2020reinforcement}. While our experiments demonstrate promising results for learning policies in simulated environments using rewards learned from both simulated and real-world videos, we have yet to show policy learning on a real robot, which we look forward to trying post-COVID.

\section*{Acknowledgments}
We would like to thank Alex Nichol, Nick Hynes, Sean Kirmani, Brent Yi, Jimmy Wu, Karl Schmeckpeper and Minttu Alakuijala for fruitful technical discussions, and Sam Toyer for invaluable help with setting up the simulated benchmark.

{
\setlength{\bibsep}{1.4pt plus 0.3ex}
\small\bibliography{main}  
}

\clearpage
\begin{appendices}

\section{Expanded Related Work}
\label{appendix:expanded_rw}
Traditional formulations of imitation learning \cite{Atkeson1997RobotLF,argall2009survey,osa2018algorithmicIL} assume access to a corpus of expert demonstration data which includes both state and action trajectories of the expert policy. In the context of third-person imitation learning, including when learning from expert agents with different embodiments, obtaining access to ground-truth actions is difficult or impossible.

\textbf{Inferring expert actions.} To address this issue, several approaches either try to infer expert actions \cite{torabi2018behavioral, edwards2018imitating, radosavovic2020state, Yang2019ImitationLF} -- for example by training an inverse dynamics model on agent interaction data~\cite{torabi2018behavioral} -- or employ forward prediction on the next state to imitate the expert without direct action supervision~\cite{pathak2018zero}. In the case of \cite{shaoconcept2robot}, a video-based action classifier trained on a large-scale human activity dataset is leveraged to provide rewards for single-task RL policies, which are then used to provide expert state-action pairs for multi-task behavior cloning. While these methods successfully address learning from observation-only demonstrations, they either do not support skill transfer to different policy embodiments at all, or they cannot take advantage of multiple embodiments in order to improve generalization to unseen policy configurations. We explicitly address these problems in this work.

\textbf{Imitation via learned reward functions.} In contrast to imitation via supervised methods, such as BCO~\cite{torabi2018behavioral}, a recent body of work~\cite{sermanet2016unsupervised, sermanet2018time, aytar2018playing, schmeckpeper2020reinforcement, mees2020adversarial, Berseth2019VisualIL} has focused on learning reward functions from expert video data and then training RL policies to maximize this reward. In \cite{sermanet2016unsupervised}, the authors combine ImageNet pre-trained visual features with an L2-norm distance reward to match policy and expert observations in a latent feature space. In their follow-up work \cite{sermanet2018time}, the reward is computed in a viewpoint-invariant representation that is self-supervised on video data. While both these methods are compelling in their use of cheap unlabeled data to learn invariant rewards, the use of a time index as a heuristic for defining weak correspondence is a constraining limitation for tasks that need to be executed at different speeds, or are not strictly monotonic (\eg have ambiguous sub-task ordering).
In \cite{aytar2018playing} a dense reward is learned via unsupervised learning on YouTube data and the authors make no assumption about time alignment. However, in their work, the expert and learned policy are executed in the same domain and embodiment, an assumption we relax in our work. Framed in a multi-task learning setting, \cite{gupta2017learning} propose training policies with morphologically different embodiments first on a similar set of proxy tasks, in order to learn a latent space to map between domains, and then sharing skills on a held-out task from one policy to another. A time-index heuristic is used to define a metric reward when performing RL training of the new task. In our work, the learned embedding finds correspondences in a fully-unsupervised fashion, without the need for such strict time alignment. In \cite{singh2019end}, a small sub-set of states is human labeled for goal success and a convolutional network is then trained to detect successful task completion from image observations, where on-policy samples are used as negatives for the classifier. By contrast, our learned embedding encodes task progress in its latent representation without the use of expensive human labels.

\textbf{Imitation via domain adaptation.} An additional category of approaches to third-person imitation learning are those that perform domain adaptation of expert observations~\cite{smith2019avid, liu2018imitation, sharma2019third, hejna2020hierarchically}. For instance, in \cite{smith2019avid} a CycleGAN~\cite{zhu2017unpaired} architecture is used to perform pixel-level image-to-image translation between policy domains, which is then used to construct a reward function for a model-based RL algorithm. A similar model-free approach is proposed in \cite{liu2018imitation}. In \cite{sharma2019third}, a generative model is used to predict robot high-level sub-goals conditioned on a third-person demonstration video, and a lower-level control policy is trained in a task-agnostic manner. Similarly, \cite{hejna2020hierarchically} uses high level task conditioning from zero-shot transfer of expert demonstrations, but they use KL matching to perform both high and low-level imitation. In contrast to these methods, the unsupervised TCC alignment in this work avoids performing explicit domain adaptation or pixel-level image translation by instead learning a robust and invariant feature space in a fully offline fashion.

Inspired by maximum entropy inverse RL~\cite{ziebart2008maximum, ziebart2010modeling} and generative adversarial networks~\cite{goodfellow2014generative}, the seminal GAIL algorithm~\cite{ho2016generative} performs distribution matching between the expert and policy's state-action occupancy via an adversarial formulation; a discriminator is trained with on-policy samples, which is then used as a reward in an RL framework. Many recent works~\cite{torabi2018generative, stadie2017third, lu2020adail, liu2019state, Sun2019ProvablyEI} build upon GAIL in order to perform observation-only imitation learning using state-only occupancy matching~\cite{torabi2018generative}, domain adaptation via domain confusion~\cite{stadie2017third, lu2020adail}, and state-alignment using a variational autoencoder next-state predictor~\cite{liu2019state}. Likewise, the algorithm proposed in \cite{mees2020adversarial} combines a metric learning loss that uses temporal video coherence to learn a robust skill representation with an entropy-regularized adversarial skill transfer loss. Finally, the authors of ~\cite{kimdomain} propose an adversarial formulation for learning across domains with dynamics, embodiment and viewpoint mismatch. In contrast to these methods, our unsupervised reward is robust to domain shift without requiring online fine-tuning or the additional complexity of dynamic reward learning.

\textbf{Reinforcement learning with demonstrations.} Recent work in offline-reinforcement learning ~\cite{schmeckpeper2020reinforcement} explicitly tackles the problem of policy embodiment and domain shift. Their method, Reinforcement Learning from Videos (RLV), uses a labelled collection of expert-policy state pairs in conjunction with adversarial training to learn an inverse dynamics model jointly optimized with the policy. In contrast, we avoid the limitation of collecting human-labeled dense state correspondences by using a self-supervised algorithm (\ie TCC~\cite{Dwibedi_2019_CVPR}) which uses cycle-consistency to automatically learn the correspondence between states of two domains. We also show that this formulation improves generalization to unseen embodiments. Since the problem setup is similar to ours, we also compare to their method as a baseline.

\section{X-MAGICAL Benchmark Details}
\label{appendix:x_magical}
In this section, we provide more information about our X-MAGICAL benchmark, including a description of the task, an overview of the different embodiments and details regarding horizons, the success metric and the environment reward. We encourage the reader to read \cite{toyer2020magical} for an in-depth description of the base MAGICAL benchmark. 

\subsection{Action and Observation Space}
\label{app:xmagical_spaces}

\begin{figure}[htbp]
    \centering
    \includegraphics[width=.5\textwidth]{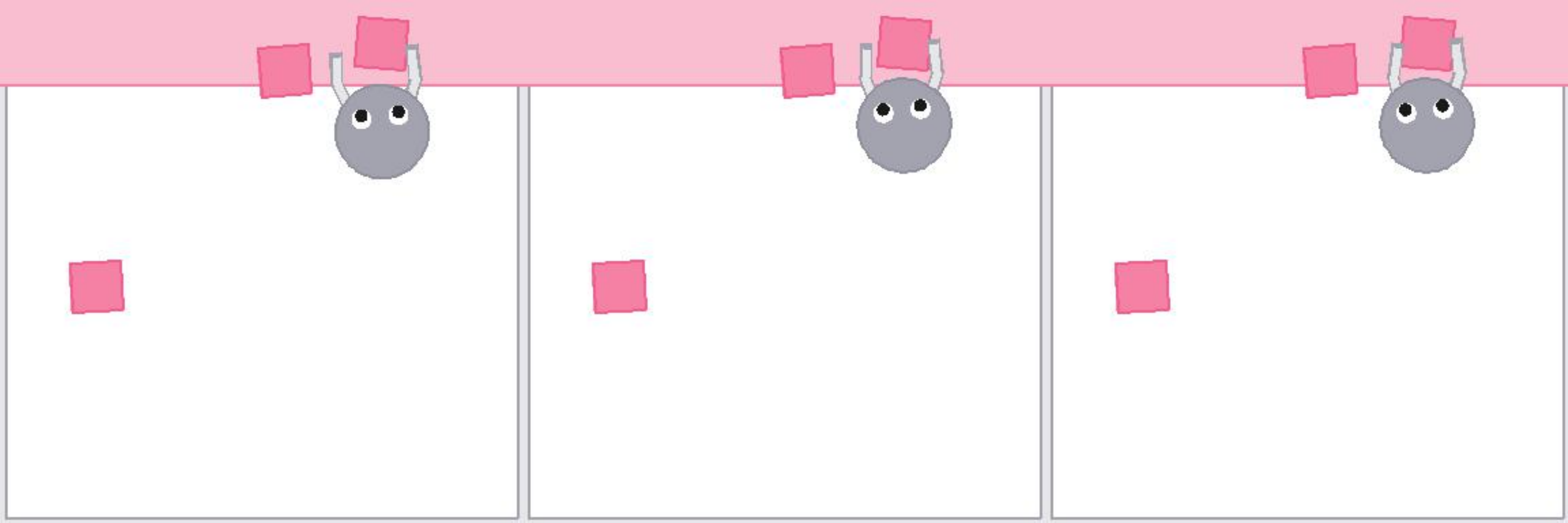}
    \caption{A stack of three image observations for the \textit{gripper} embodiment. }
    \label{fig:allo_gripper}
\end{figure}

We use a continuous action space for our \textit{Sweeping} task. All embodiments have a 2D action space with the exception of the \textit{gripper} agent which has an additional degree of freedom to open and close its arms. The first degree of freedom is for longitudinal movement (forward/backward), the second degree of freedom is for angular movement (left and right rotation) and the third degree of freedom, if applicable, is a gripping action (push fingers closed/allow fingers to open).

x-MAGICAL provides both state and image-based observations. The state observations are used as input to the RL policies whereas the pixel observations are used by the pretrained encoder to generate rewards. The state vector contains the $(x,y)$ position of the agent, $(\cos{\theta},\sin{\theta})$ where $\theta$ is the agent's 2D orientation, and for each of the three debris: its $(x,y)$ position, its distance to the agent and its distance to the goal zone. For the RGB image, we employ an allocentric, top-down perspective (Figure~\ref{fig:allo_gripper}) with full view of the workspace. Similar to MAGICAL, we use an $8$Hz control rate, thus with a frame stacking value of $3$, this corresponds to roughly $0.3$ seconds of interaction.

\subsection{Detailed Task and Embodiment Description}

In the \textit{Sweeping} task, the robotic agent must push 3 debris blocks to the pink zone at the top of the workspace. The agent's position is constrained to always spawn below the debris. Both the agent's position and the debris positions are randomized at every environment reset. Specifically, we sample the same y-coordinate for all three debris, then randomly space them out from each other (different x-coordinate). The horizon for the \textit{longstick} agent is $H = 50$ time steps since it can solve the task much faster than the other embodiments thanks to its morphology. The horizon for all other embodiments is $H = 100$. The ground-truth environment reward is defined as $\frac{1}{3} \cdot \sum_{i=1}^3 \mathbbm{1}\{d_i \in G\}$, \ie the fraction of total debris present inside the goal zone $G$.

\begin{figure}[htbp]
    \centering
    \includegraphics[width=\textwidth]{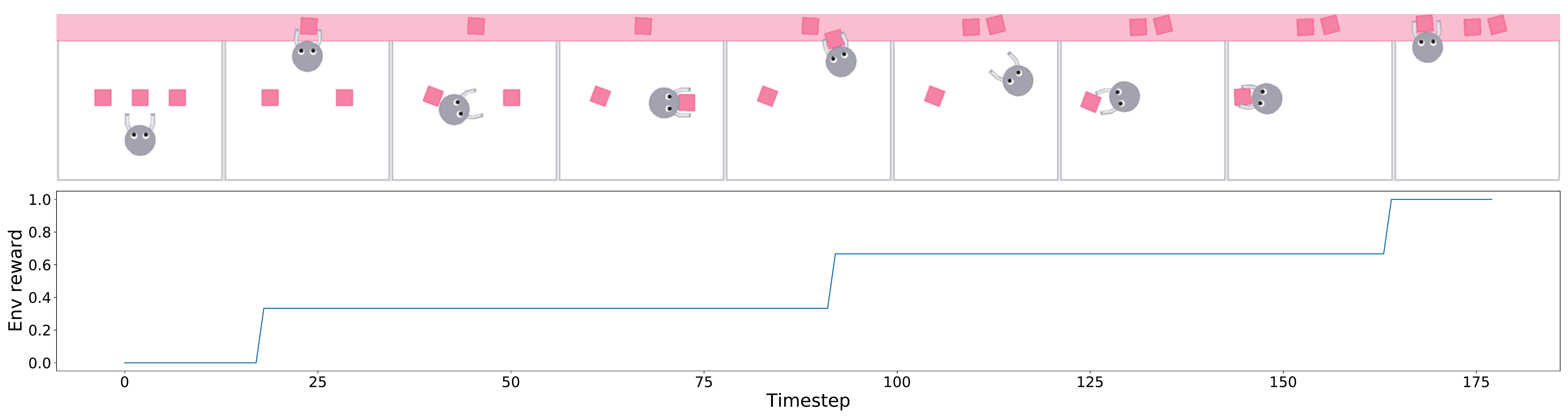}
    \includegraphics[width=\textwidth]{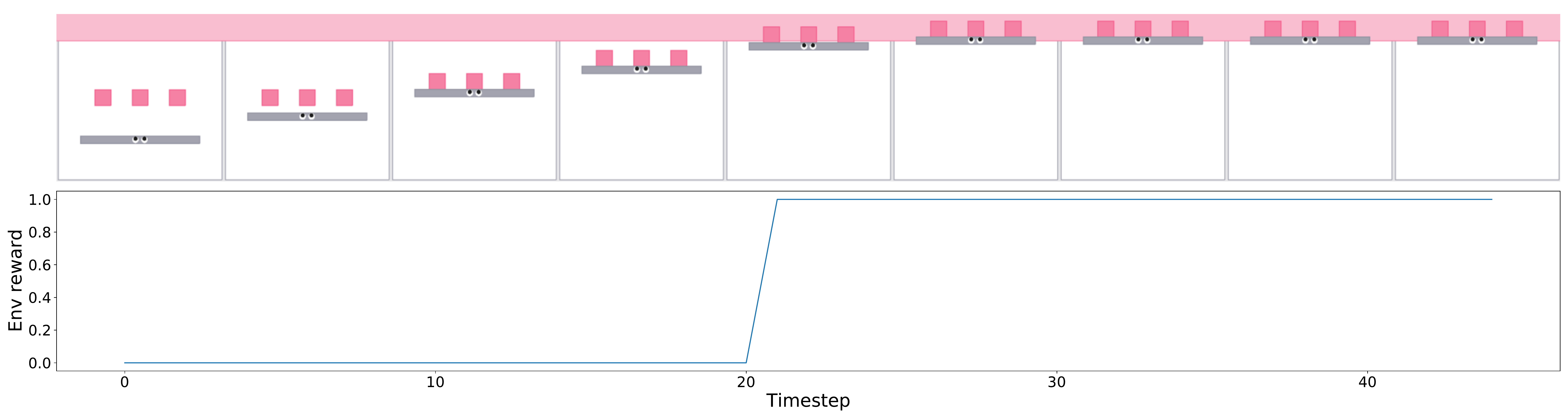}
    \includegraphics[width=\textwidth]{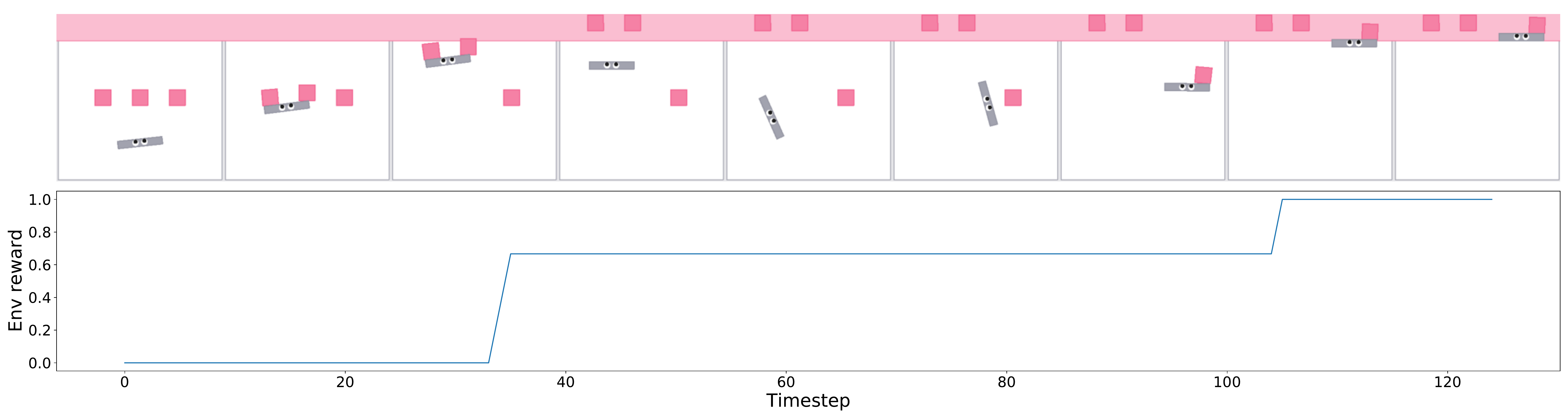}
    \includegraphics[width=\textwidth]{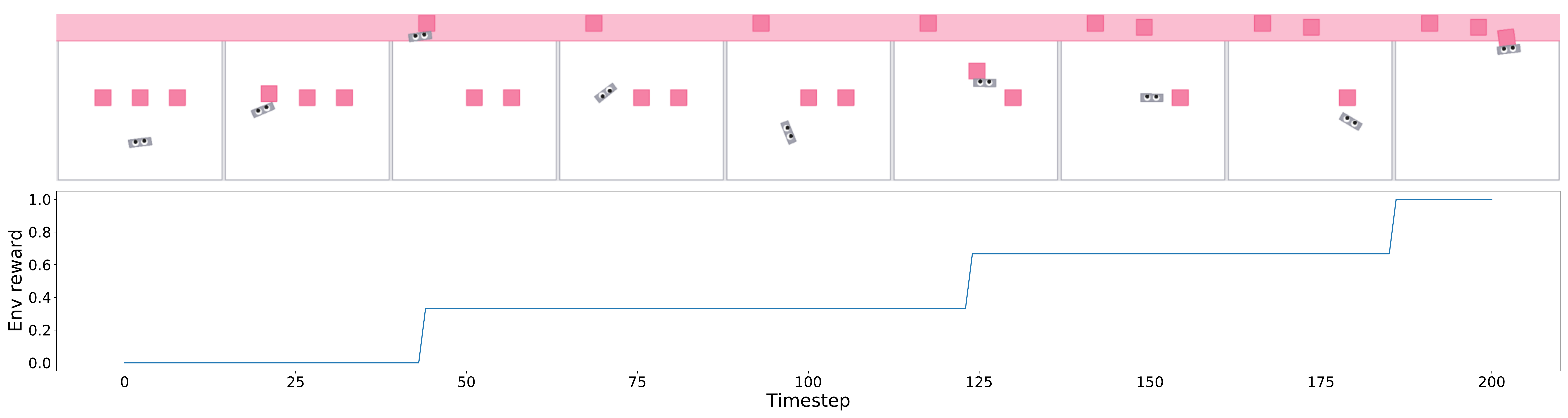}
    \caption{A film strip of the \textit{gripper}, \textit{longstick}, \textit{mediumstick} and \textit{shortstick} embodiments (vertical order) solving the Sweeping task with a corresponding visualization of the ground-truth environment reward.}
    \label{fig:demo_strip}
\end{figure}

In Figure~\ref{fig:demo_strip}, we provide a film strip demonstration of each embodiment solving the \textit{Sweeping} task with a plot of the environment reward as a function of time. For this visualization specifically, we manually teleoperate each agent and disable the environment horizon limit.

\subsection{Demonstration Data}

To collect demonstration data for each embodiment, we train an oracle policy with SAC using the ground-truth environment reward. We then rollout the policy in the environment, discarding any potentially unsuccessful demonstrations, until we are left with $1000$ demonstrations per embodiment. A comprehensive overview of the hyperparameters used for reinforcement learning are detailed in Table~\ref{tab:hp-rl} in Appendix~\ref{app:hparams_pl}.

\section{X-REAL: A Real-World Cross-Embodiment Dataset}
\label{appendix:x_real}
\begin{table}
\centering
\begin{tabular}{l|c}
\toprule
\textbf{Embodiment} & \textbf{Demo Length Stats} \\
& \textbf{(seconds)} \\
\midrule
\midrule
1 Hand 5 Fingers      & $10.8 \pm 1.8$ \\
2 Hands 2 Fingers     & $11.8 \pm 1.2$ \\
1 Hand 2 Fingers      & $14.9 \pm 1.4$ \\
Ski Gloves              & $20.0 \pm 3.9$ \\
Kitchen Tongs           & $21.1 \pm 3.1$ \\
Lobster Hands (costume) & $21.3 \pm 3.8$ \\
Tweezers                & $27.8 \pm 2.3$ \\
RMS Grabber Reacher     & $29.9 \pm 3.2$ \\
Irwin Quick-Grip Clamps & $38.1 \pm 9.2$ \\
\bottomrule
\end{tabular}
\caption{Mean $\pm$ Std. Dev. of Demonstrations Lengths for Different X-REAL Embodiments}
\label{tab:stats_real}
\vspace{-2em}
\end{table}

\begin{figure*}[htbp]
    \centering
    \includegraphics[width=\textwidth]{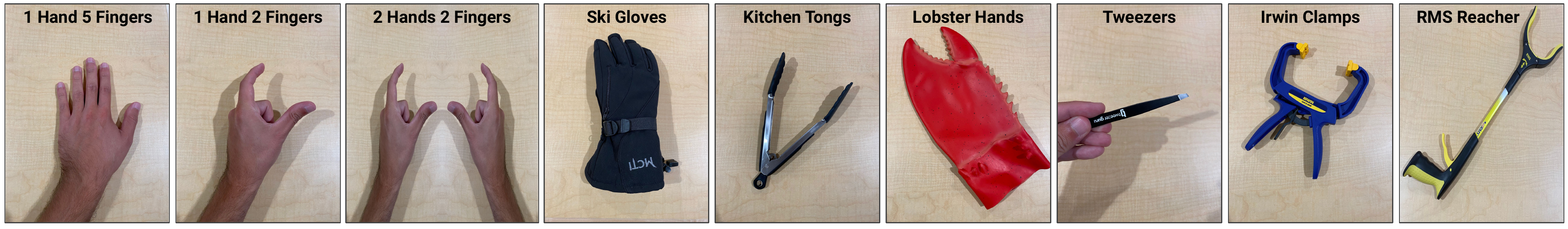}
    \caption{Embodiments in the X-REAL dataset, ordered by their appearance in Table \ref{tab:stats_real}.}
    \label{fig:x_real_embs}
    \vspace{-1em}
\end{figure*}

To test reward learning in the real world on more challenging manipulation tasks, we collect a real-world dataset named X-REAL (Cross-embodiment Real-world demonstrations), which contains 93 demonstration videos of different embodiments (manifested as different manipulator end-effectors) solving the same manipulation task in the real-world: \textit{transferring five pens to two cups consecutively}. This is a multi-step manipulation task where the pens on the table need to be lifted to one cup and then moved again to a separate cup. The different end-effectors consist in a human hand as well as six tools purchased from Amazon and displayed in Figure \ref{fig:x_real_embs}. In contrast to the dataset used in Section~\ref{subsection:real-cross-embodiment}, there is visual diversity among the different end-effectors in X-REAL, and there is also significant variation in how the task is solved and how long it takes to solve the task. Some end-effectors (\eg tweezers) can only move one pen at a time, while others (\eg human hand with five fingers) can move all pens at once. Additionally, the demonstrations are not collected in a constrained fashion that tries to mimic the robot. We report the mean and standard deviation of demonstration lengths for each embodiment in Table~\ref{tab:stats_real}. Note the variation in demo lengths across different embodiments.


\subsection{Data Collection \& Hardware Setup}

\begin{wrapfigure}{r}{0.35\textwidth}
  \vspace{-1.8em}
  \begin{center}
    \includegraphics[width=0.35\textwidth]{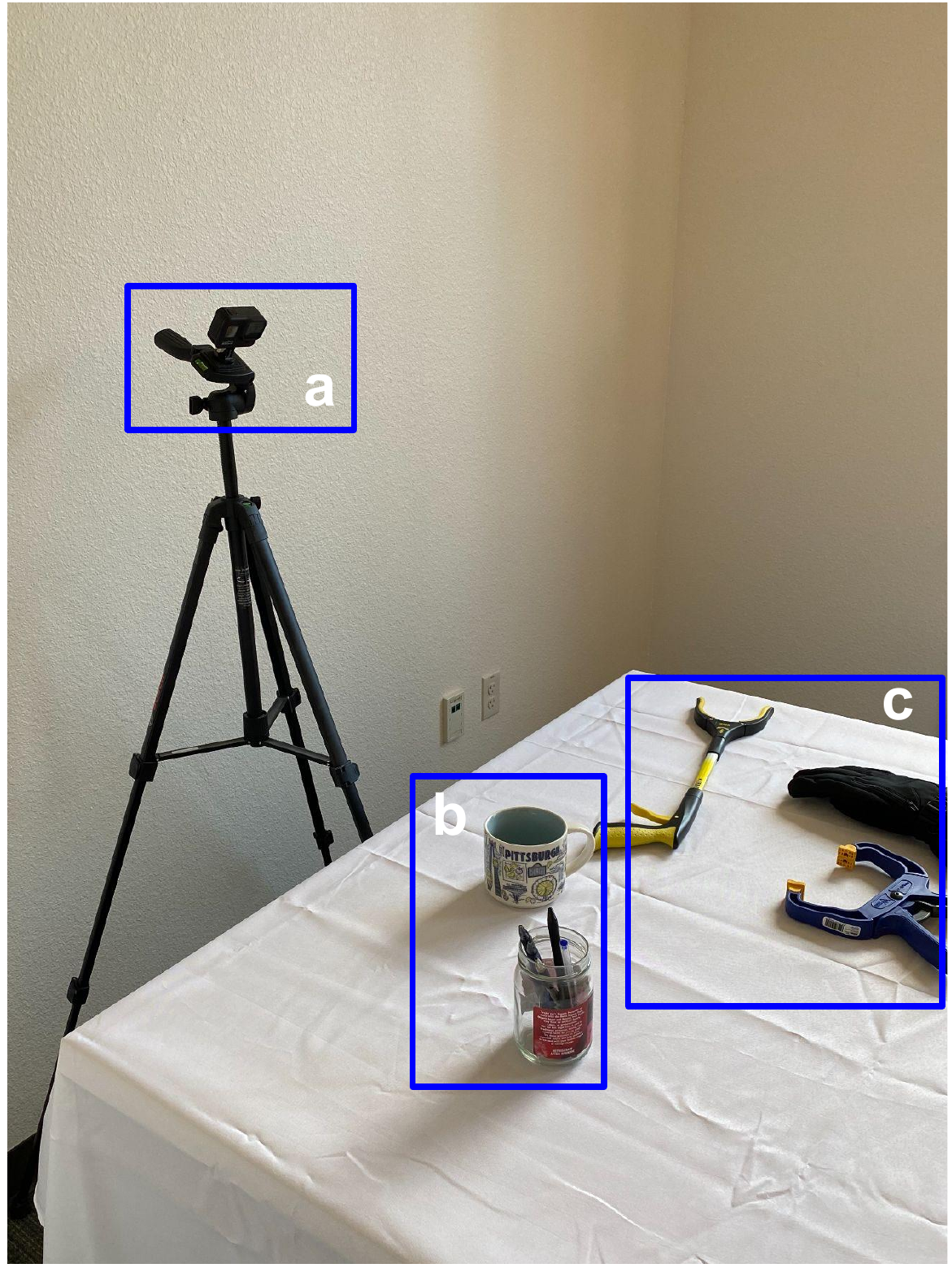}
  \end{center}
  \vspace{-0.5em}
  \caption{\textbf{Hardware setup.} a) Tripod-mounted GoPro Hero8, b) Cup, Mug and Pens from Task c) Example end-effectors used.}
  \label{fig:hardware-setup}
  \vspace{-2em}
\end{wrapfigure}

The hardware and data collection setup is shown in Figure \ref{fig:hardware-setup}. We use a GoPro Hero8 mounted on a tripod to record the demonstrations and use voice commands to efficiently start and stop the video recordings. The Hero8 records RGB images with a resolution of $1920 \times 1080$ at $30$ frames per second. 


\subsection{Reward Learning from \textbf{Real-world Multi-Cross-Embodiment} Demonstrations}
\label{subsection:real-world-demonstrations}

\begin{figure*}[htbp]
\centering
    \includegraphics[width=\textwidth]{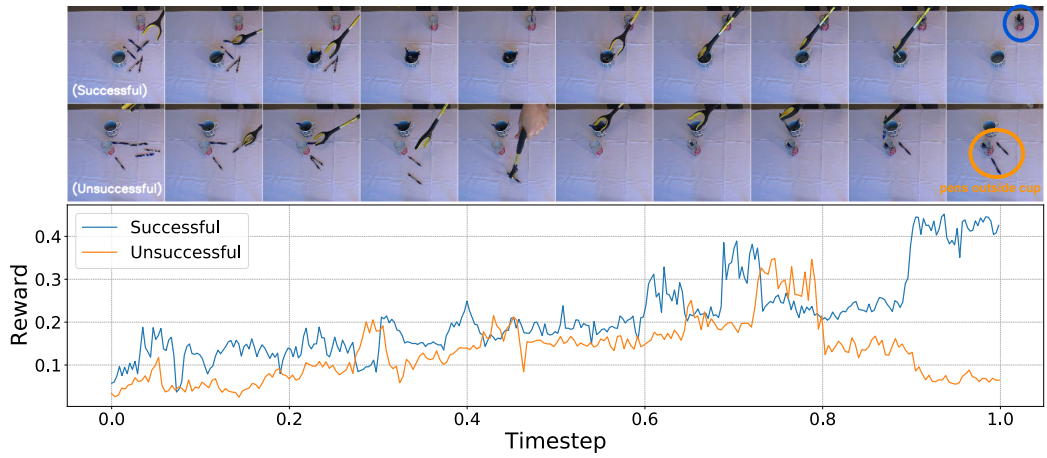}
    \includegraphics[width=\textwidth]{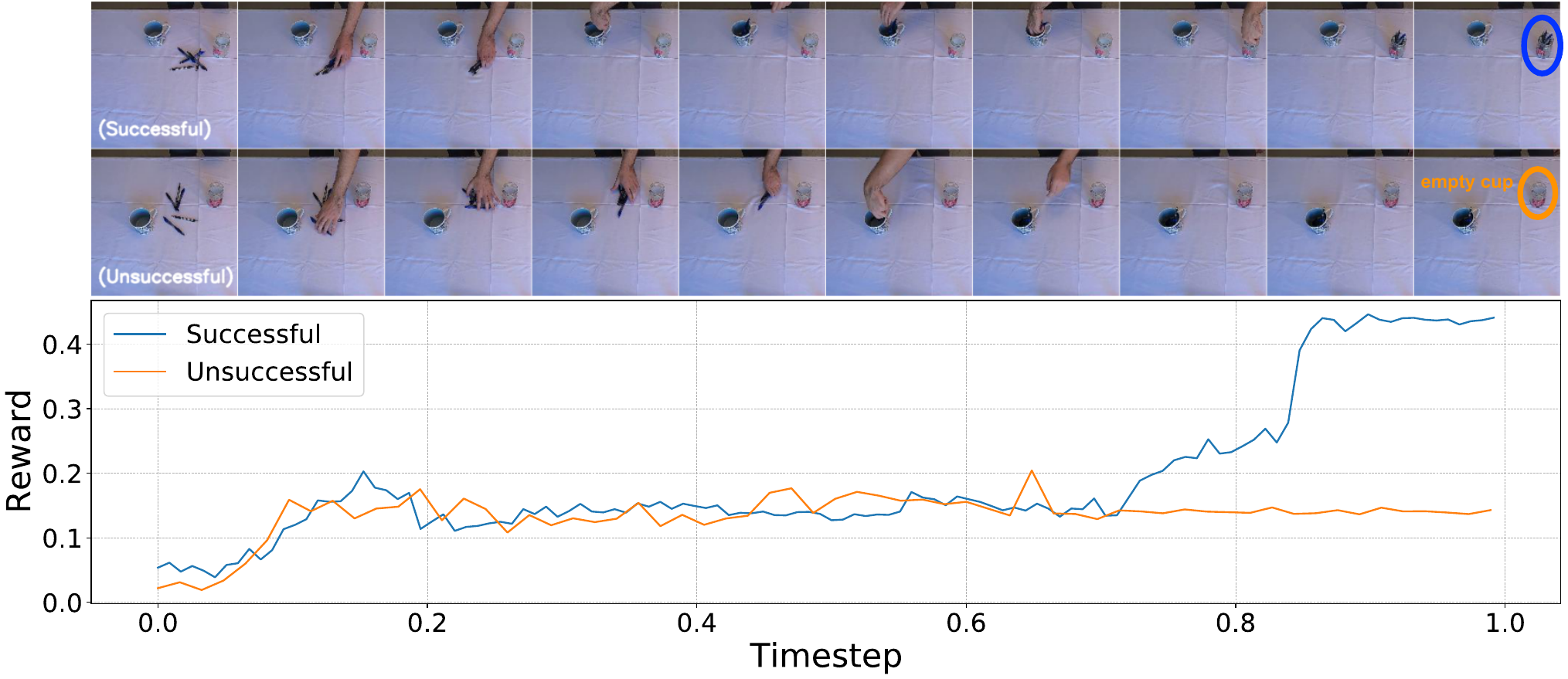}
    \caption{\textbf{\textit{X-REAL ``move pens to mug then cup'' task:}} Visualizing our learned XIRL reward function on successful and unsuccessful demonstrations for the \textit{RMS Grabber Reacher} (\textit{top}) and \textit{1 Hand 5 Fingers} (\textit{bottom}) embodiments.}
    \label{fig:rms_reward}
\end{figure*}

In this section, we learn reward functions from videos in X-REAL, and demonstrate that our method is capable of handling the visual complexities of the real world without requiring annotations of end-effectors, objects, or their states. We train the encoder on all embodiments in the training set and present examples of the learned XIRL rewards on video demonstrations from the validation set in Figure~\ref{fig:rms_reward}. Specifically, we visualize two embodiments: the \textit{RMS Grabber Reacher} (top row) and the human \textit{1 Hand 5 Fingers} (bottom row) and for each embodiment, we show both a successful and unsuccessful trajectory.

In the top row, both the successful and unsuccessful demonstrations follow a similar trajectory at the start of the task execution. The successful one nets a high reward for placing the pens consecutively into the mug then into the glass cup, while the unsuccessful one obtains a low reward because it drops the pens outside the glass cup towards the end of the execution. In the bottom row, for the \textit{1 Hand 5 Fingers} embodiment, we observe that not completing the task and more specifically, leaving the pens in the first cup, generates a reward that is roughly half (image row 2, plot orange curve) the one achieved by a successful execution (image row 1, plot blue curve). These results are encouraging -- they show that our learned encoder can represent fine-grained visual differences relevant to the task. Additionally, the training process for this visual reward did not require any additional environment instrumentation (apart from a camera), a desirable property for scaling to more complex, multi-step manipulation tasks.

\clearpage

\section{Additional Experimental Details}
\label{appendix:experimental}
Our codebase is implemented in PyTorch \cite{NEURIPS2019_9015}. Experiments were performed on a desktop machine with an AMD Ryzen 7 2700X CPU (8 Cores/16 Threads, 3.7GHz base clock), 32GB RAM, and a single NVIDIA GeForce RTX 2080 Ti GPU.

\subsection{Representation Learning}
Each representation learning run -- specifically training and evaluating a representation and computing the final goal embedding vector -- took an average of $25$-$30$ minutes of wall clock time.

\subsubsection{Additional Baseline Details}
\label{app:baseline_details}

In this section, we provide a comprehensive overview of the baselines briefly described in Section~\ref{sec:baselines}.

\textbf{ImageNet}: We use an ImageNet pre-trained ResNet-18 with no additional training, \ie we load the pre-trained weights, discard the classification head, and use the $512$-dimensional embedding space from the penultimate layer.

\begin{wraptable}{r}{5.5cm}
\scriptsize
\centering
\begin{tabular}{l|r}
Parameter        & Setting \\
\midrule
Total train iterations & $6k$ \\
Number of frames & $15$ \\
Batch size & $4$ \\
Embedding size & $32$ \\
Unit normalize embeddings & False \\
Negative window size & $5$ \\
\end{tabular}\\
\caption{\label{table:goal_cls} Hyperparameters used for the \textbf{Goal Classifier} baseline.}
\vspace{-1em}
\end{wraptable} 

\textbf{Goal Classifier}: A common strategy for using a learned visual reward function is to train a goal frame classifier \cite{vecerik2019practical} on a binary classification task where the last frame of all the demonstrations is considered positive and all the others are considered negatives. Since demonstration sequences do not necessarily end exactly when the task is solved, we randomly sample the negatives from frames that are at least $5$ indices away from the final frame. This was chosen heuristically by examining the dataset. Note that unlike \cite{vecerik2019practical}, we do not have access to any unsuccessful trajectories in the dataset.

\begin{wraptable}{r}{5.5cm}
\scriptsize
\centering
\begin{tabular}{l|r}
Parameter        & Setting \\
\hline
Total train iterations & $8k$ \\
Number of frames & $15$ \\
Batch size & $4$ \\
Embedding size & $32$ \\
Unit normalize embeddings & True \\
Embedding temperature & $0.1$ \\
\end{tabular}\\
\caption{\label{table:lifs} Hyperparameters used for the \textbf{LIFS} baseline.}
\vspace{-1em}
\end{wraptable} 

\textbf{LIFS}: We re-implement the approach from \cite{gupta2017learning}, which learns a feature space that is invariant to different embodiments using a loss function that encourages \textit{corresponding} pairs of embodiment states across demonstrations to be close in the embedding space. We use the time-based alignment method described in their paper to find these corresponding pairs which assumes each embodiment performs the task at the same rate. To prevent the embeddings from collapsing to a constant value, they use an additional reconstruction loss to encourage the network decoders to preserve as much domain-invariant information as possible. We found early stopping to be crucial in preventing LIFS from collapsing to trivial embeddings. For this baseline, we also randomly sample $N$ evenly-spaced frames from a video to construct each mini-batch.

\begin{wraptable}{r}{5.5cm}
\scriptsize
\centering
\begin{tabular}{l|r}
Parameter        & Setting \\
\midrule
Total train iterations & $4k$ \\
Number of frames & $20$ \\
Batch size & $4$ \\
Embedding size & $32$ \\
Unit normalize embeddings & True \\
Embedding temperature & learned \\
Positive window & $1$ \\
Negative Window & $4$ \\
\end{tabular}\\
\caption{\label{table:tcn} Hyperparameters used for the \textbf{TCN} baseline.}
\vspace{-2em}
\end{wraptable}

\textbf{TCN}: We re-implement the single-view variant of Time-Contrastive Network (TCN) \cite{sermanet2018time} with positive and negative frame windows of $1$ and $4$ respectively. Different from \cite{sermanet2018time}, we do not use a time-indexed reward which is not applicable to agents with different embodiments. Like XIRL, we use the negative distance to the average goal embedding as the reward. For this baseline, we sample a contiguous chunk of $N$ frames from a video to construct a mini-batch.

\subsubsection{Data Augmentation \& Preprocessing}
We apply data augmentation during training using the \textbf{Albumentations} library \cite{alb2018aug}. Concretely, this involves the following transformations:
\begin{itemize}
    \item \texttt{\href{https://albumentations.ai/docs/api_reference/augmentations/crops/transforms/\#albumentations.augmentations.crops.transforms.RandomResizedCrop}{RandomResizedCrop}}: This transformation takes a random crop from the original image, then resizes it to a final output height and width. The lower and upper bounds for the random crop area are set to $[0.6, 1.0]$, the lower and upper bounds for the random aspect ratio of the crop are set to $[0.75, 1.33]$ and the final output size is set to $224 \times 224$. We apply this transformation with a probability of $1.0$ and denote it by $C$.
    \item \texttt{\href{https://albumentations.ai/docs/api_reference/augmentations/transforms/\#albumentations.augmentations.transforms.ColorJitter}{ColorJitter}}: This transformation varies the brightness, contrast, saturation and hue of an input image. Our parameters for these respective changes are $0.4$, $0.4$, $0.1$ and $0.1$. We apply it with a probability of $0.8$ and denote it by $J$.
    \item \texttt{\href{https://albumentations.ai/docs/api_reference/augmentations/transforms/\#albumentations.augmentations.transforms.ToGray}{ToGray}}: This transformation converts an RGB image into a grayscale one. We apply it with a probability of $0.2$ and denote it by $G$.
    \item \texttt{\href{https://albumentations.ai/docs/api_reference/augmentations/transforms/\#albumentations.augmentations.transforms.GaussianBlur}{GaussianBlur}}: This transformation blurs the input image with a Gaussian filter. We use a fixed kernel size of 13 and a standard deviation randomly sampled from the range $[1.0, 2.0]$. We apply it with a probability of $0.2$ and denote it by $B$.
    \item \texttt{\href{https://albumentations.ai/docs/api_reference/pytorch/transforms/\#albumentations.pytorch.transforms.ToTensor}{Normalize}}: Lastly, we divide the pixel values by 255 to scale their range to $[0, 1]$. We apply it with a probability of $1.0$ and denote it by $N$.
\end{itemize}

We apply the same randomly sampled transformations to all the sampled frames from the same video. However, independently sampled transformations are applied for each such frame stack in the mini-batch. Note that the order of transformation matters, \ie we apply the following composed transform:  $N \circ B \circ G \circ J \circ C$.

\subsubsection{Training and Evaluation}
All our representations are trained using an ADAM optimizer with $\beta_1 = 0.99$, $\beta_2 = 0.999$ and weight decay of $10^{-5}$. While the representations are evaluated on the downstream policy learning performance, we also compute the following quantitative metrics and qualitative results on the train and validation sets to diagnose our representations:
\begin{itemize}
    \item \textbf{Kendall's Tau}: A metric ranging from $[-1, 1]$ that measures how well-aligned two sequences are in time. We refer the reader to \cite{Dwibedi_2019_CVPR} for a more in-depth explanation.
    \item \textbf{Nearest-Neighbor Alignment Video}: We randomly select one demonstration as a reference video. We use nearest-neighbor in the embedding space to align a test video with the reference video. See Appendix~\ref{appendix:qualitative_results} and the supplemental video for example visualizations for both \textit{same} and \textit{cross}-embodiment settings. These videos highlight how well the embedding space encodes the task progress across different embodiments. 
\end{itemize}

\subsubsection{Other Hyperparameters}
For a comprehensive list of hyperparameters used for representation learning on the X-MAGICAL environment, the Puck Pushing environment, and X-REAL (Appendix~\ref{appendix:x_real}), see Appendix~\ref{app:hparams_repr}.

\subsection{Policy Learning}
The SAC~\cite{haarnoja2018soft} implementation we use is based off of~\cite{yarats2020pytorch_sac}. Each run on the X-MAGICAL benchmark, \ie the training and evaluation of a specific reward learning method on a specific embodiment with a single seed, took an average of $00$h$27$m, $01$h$42$m, $03$h$56$m and $03$h$56$m wall clock times for \textit{long-stick}, \textit{medium-stick}, \textit{short-stick} and \textit{gripper}, respectively. Note the difference in run times is due to the fact that each embodiment is trained for a different number of total training steps since each converges at different rates: $75$k, $225$k, $500$k and $500$k for \textit{long-stick}, \textit{medium-stick}, \textit{short-stick} and \textit{gripper} respectively. For the Puck Pushing experiments in Section~\ref{subsection:real-cross-embodiment}, each run took an average wall clock time of 01h25m for $200$k timesteps. Note that run times for both the above environments are recorded while performing up to 5 seed runs in parallel.

\subsubsection{Soft-Actor Critic Architecture}
We use clipped double Q-learning~\cite{hasselt2015doubledqn, fujimoto2018td3} for the critic, where each $Q$-function is parameterized by a 3-layer multi-layer perceptron (MLP) with \texttt{ReLU} activations. The actor is implemented as a \texttt{tanh}-diagonal-Gaussian, and is also parameterized by a 3-layer MLP which outputs mean and covariance. Both actor and critic MLPs have a hidden size of $1024$ -- the weights are initialized with orthogonal~\cite{saxe2013ortho} initialization, while the biases are initialized to zero.

\subsection{Policy Input}
As mentioned in Section~\ref{subsection:x-magical} and Appendix~\ref{app:xmagical_spaces}, we construct our observational inputs by stacking $3$ consecutive state vectors. The input to the policy is thus a flattened vector in $\mathbb{R}^{48}$. For the \textit{Puck Pushing} environment described in Section~\ref{subsection:real-cross-embodiment}, we construct the observational input by stacking $3$ consecutive state vectors containing the 3D Cartesian coordinates of the robot end-effector and the planar 2D coordinates of the puck. The input to the policy is thus a flattened vector in $\mathbb{R}^{15}$.

\subsubsection{Training and Evaluation Setup}
We first collect $5000$ seed observations with a uniform random policy, after which we sample actions using the SAC policy. We then perform one gradient update every time we receive a new environment observation. When evaluating our agent every $5000$ steps, we take the mean policy output (\ie no sampling) and average the final success rate over $50$ evaluation episodes.

\subsection{Other Hyperparameters}
For a comprehensive list of hyperparameters used for policy learning on the X-MAGICAL and Puck Pushing environments, see Appendix~\ref{app:hparams_pl}.

\section{Additional Experiments}
\label{appendix:additional_experiments}
\subsection{XIRL vs. Ground-truth Environment Reward}

\begin{figure}[htbp]
    \centering
    \includegraphics[width=\textwidth]{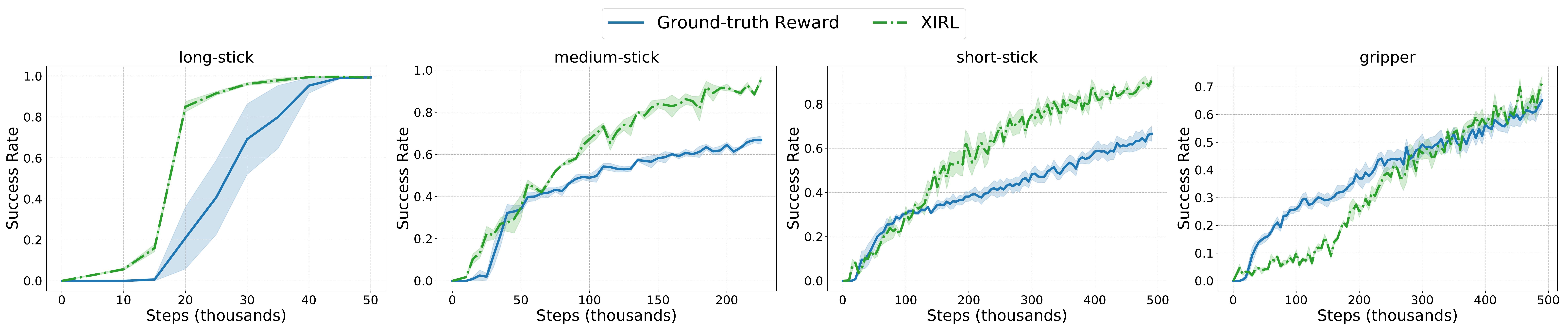}
    \includegraphics[width=\textwidth]{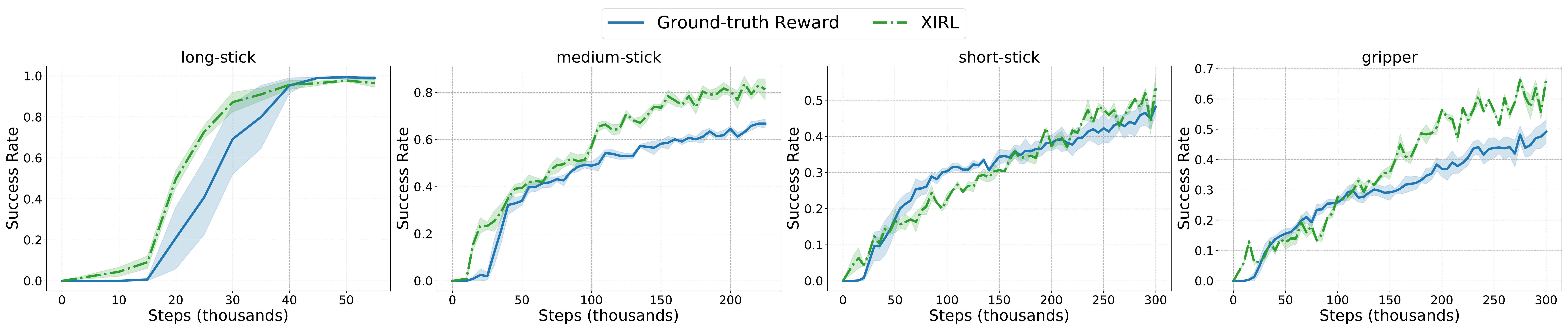}
    \caption{Comparison of XIRL with the ground-truth environment reward in the \textbf{\textit{same-embodiment setting}} (top) and the \textbf{\textit{cross-embodiment setting}} (bottom}
    \label{fig:tcc_vs_env}
\end{figure}

In this section, we compare the performance of our method XIRL against the ground-truth environment reward on the same-embodiment experiment from Section~\ref{subsection:same-embodiment} and the cross-embodiment experiment from Section~\ref{subsection:cross-embodiment}.
We observe that in both same-embodiment (Figure~\ref{fig:tcc_vs_env}, \textbf{top}) and cross-embodiment (Figure~\ref{fig:tcc_vs_env}, \textbf{bottom}) settings across multiple embodiments, XIRL is either more or just as sample-efficient as the environment reward. This highlights XIRL's ability to provide denser reward information via encoding task progress, as opposed to the sparser ground-truth environment reward, which was shown in Figure~\ref{fig:qual_reward_plots_magical}(a).

\subsection{XIRL vs. SimCLR}

In this section, we compare the performance of XIRL against SimCLR \cite{chen2020simple}, a constrative pretraining technique that has exhibited SOTA self-supervised performance on ImageNet. We implement two SimCLR baselines: (a) \textit{SimCLR} is a ResNet18 trained on the x-MAGICAL demonstration dataset with the constrastive pretraining pipeline described in \cite{chen2020simple}, and (b) \textit{SimCLR ImageNet} is a ResNet18 pretrained on ImageNet with SimCLR, with no further pretraining on x-MAGICAL. Below, we present results for the \textit{longstick} and \textit{mediumstick} embodiments of the x-MAGICAL benchmark:

\begin{figure}[htbp]
    \centering
    \includegraphics[width=.45\textwidth]{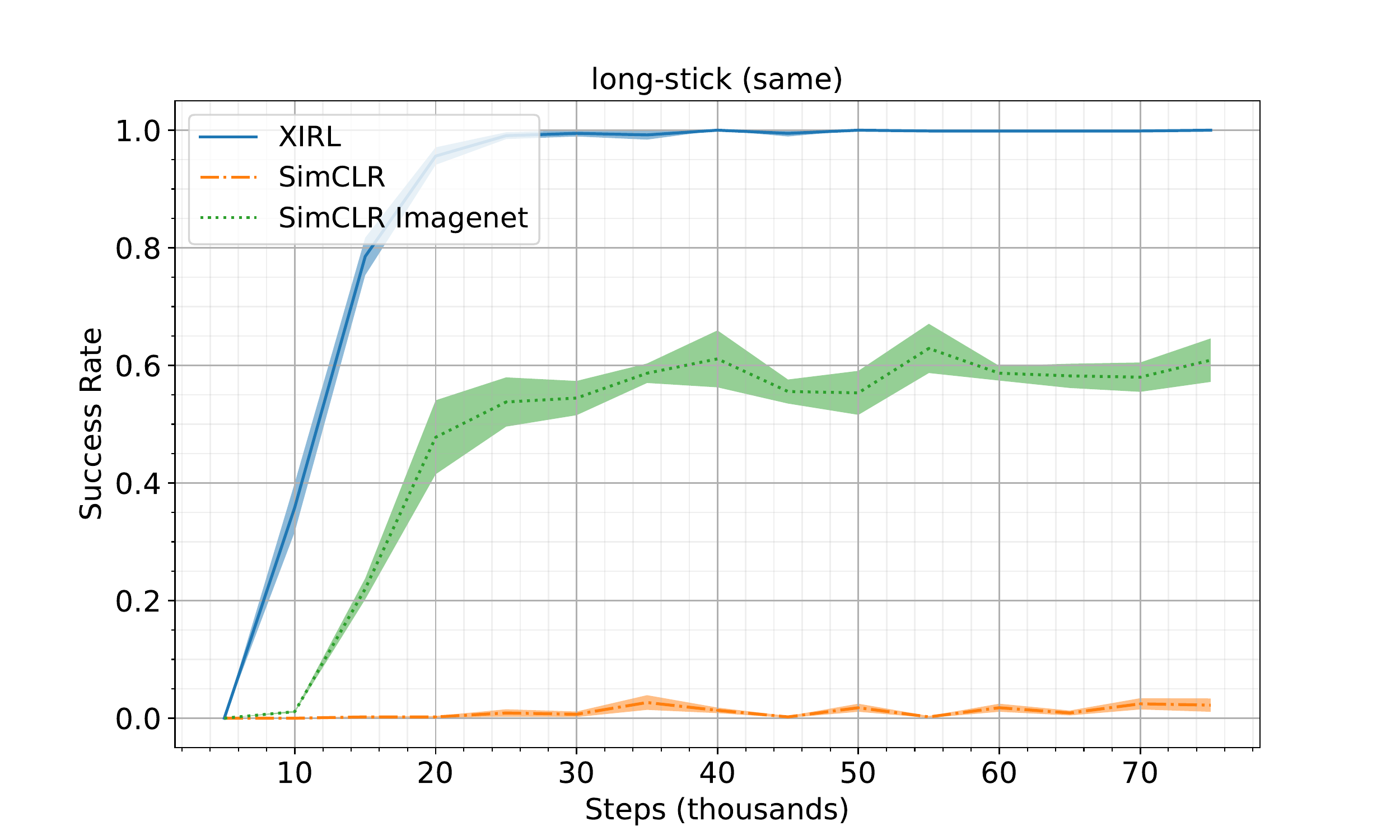}
    \includegraphics[width=.45\textwidth]{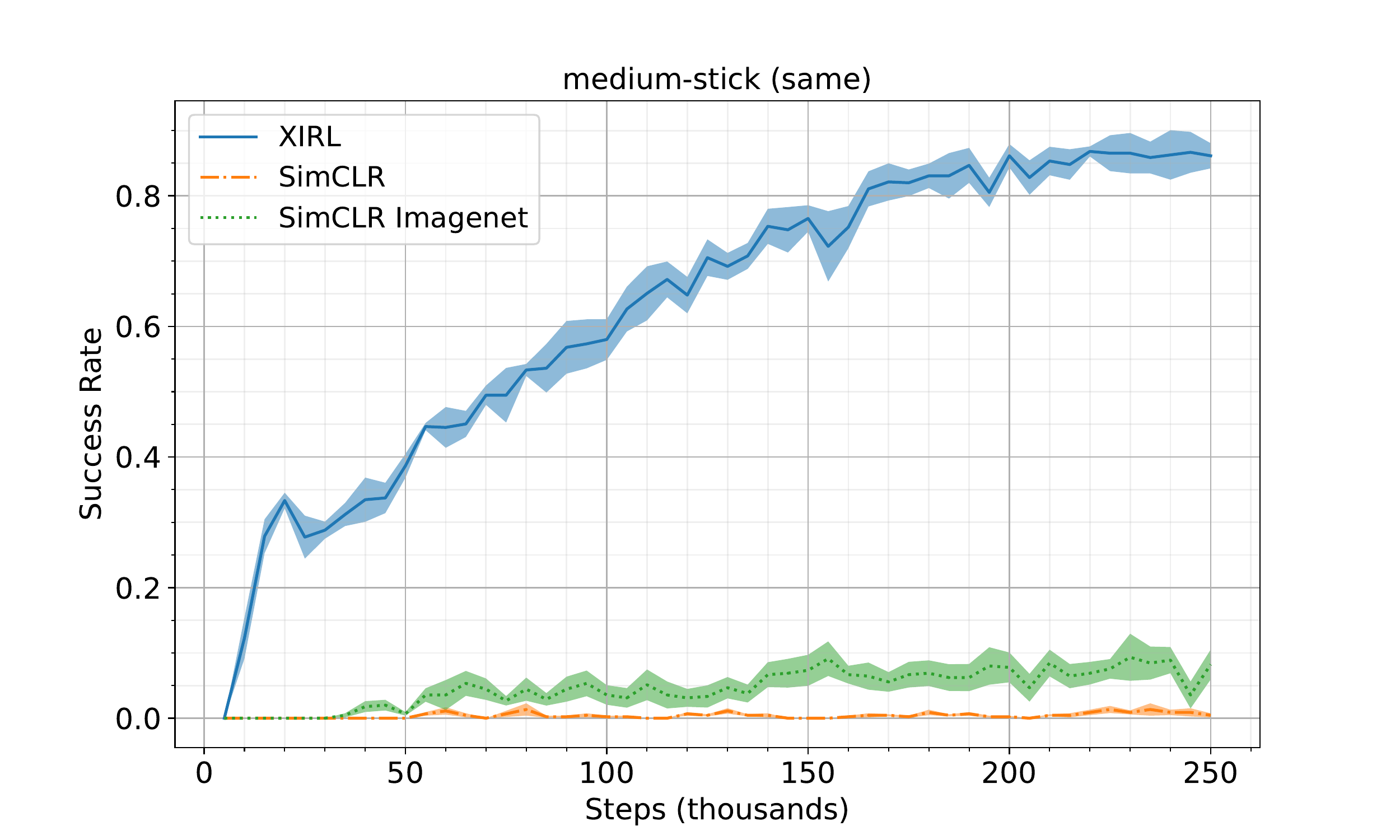}
    \caption{Comparison of XIRL \& SimCLR in the \textbf{\textit{same-embodiment setting}}.}
    \label{fig:tcc_vs_simclr_same}
    \vspace{-1em}
\end{figure}

\begin{figure}[htbp]
    \centering
    \includegraphics[width=.45\textwidth]{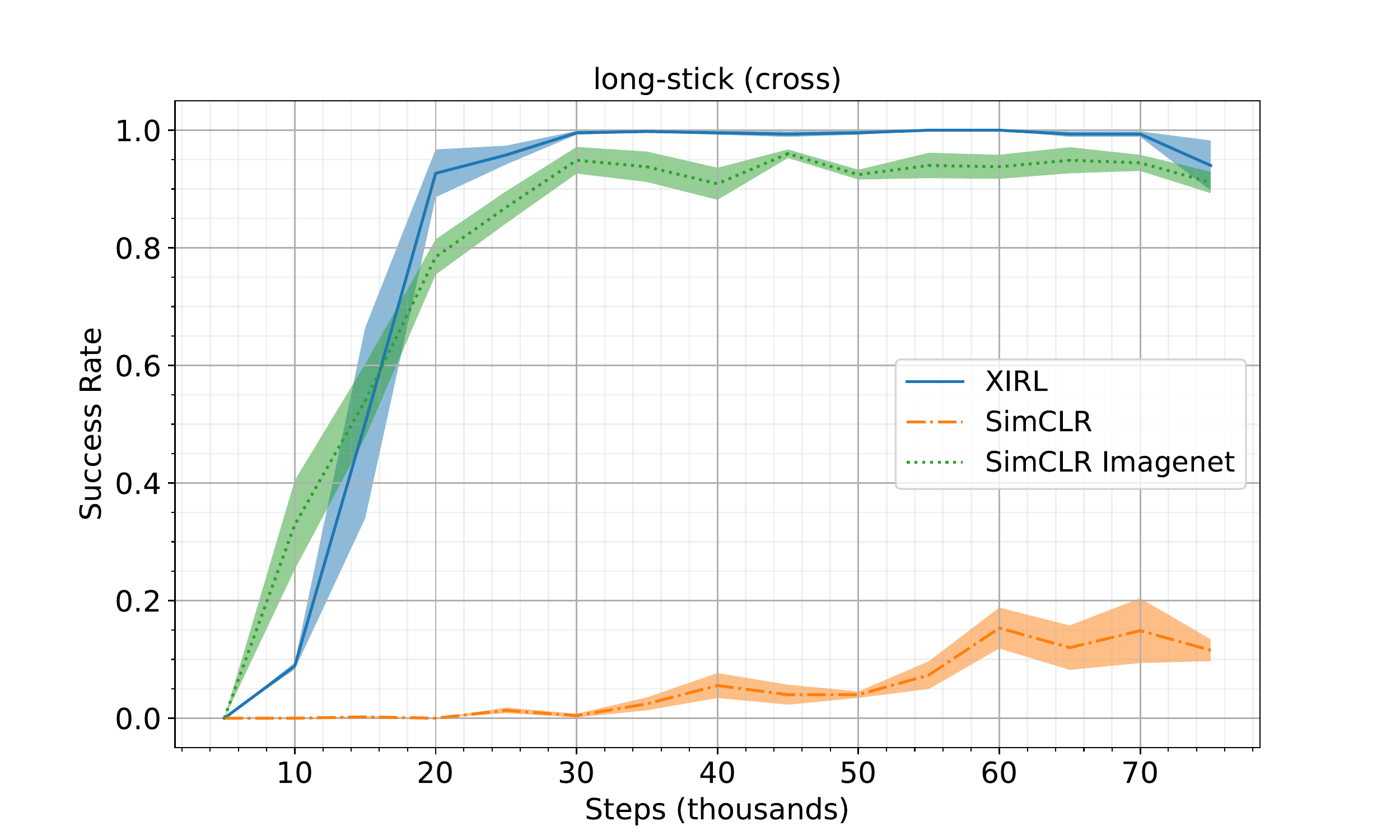}
    \includegraphics[width=.45\textwidth]{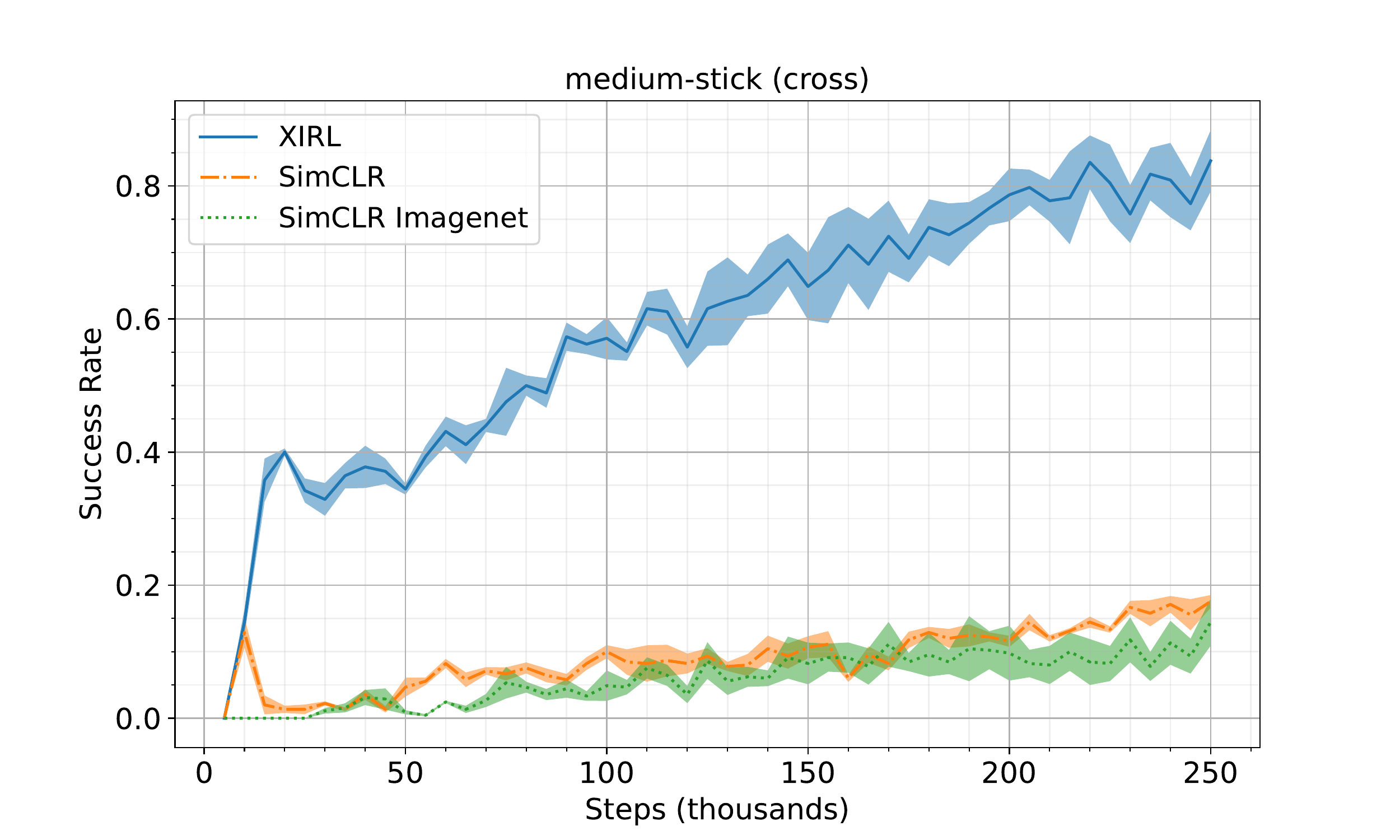}
    \caption{Comparison of XIRL \& SimCLR in the \textbf{\textit{cross-embodiment setting}}.}
    \label{fig:tcc_vs_simclr_cross}
\end{figure}

We find that the SimCLR objective performs poorly when trained solely on the x-MAGICAL dataset, whereas the SimCLR baseline pertained on ImageNet (without finetuning on x-MAGICAL) does much better on the longstick embodiment (which is easier to solve). For the mediumstick embodiment, both perform poorly. XIRL, shown in blue, performs significantly better. This highlights that overall, both visual pretraining on cross-embodiment demonstrations and the inductive biases offered by the TCC loss are required to obtain good performance on downstream RL tasks.

\section{Qualitative Results}
\label{appendix:qualitative_results}
\subsection{t-SNE Visualizations}
\begin{figure*}[htbp]
\centering
    \includegraphics[width=\textwidth]{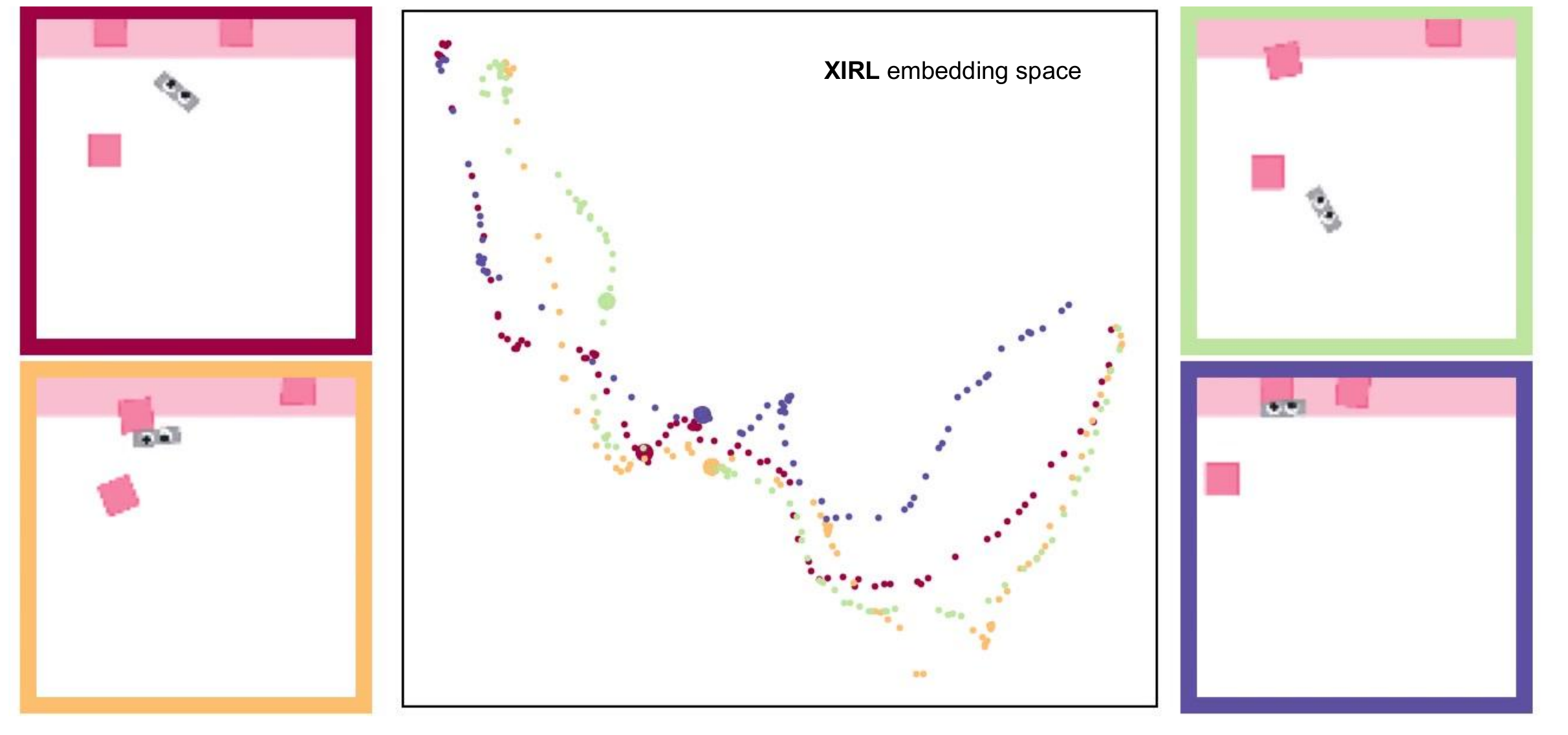}
    \includegraphics[width=\textwidth]{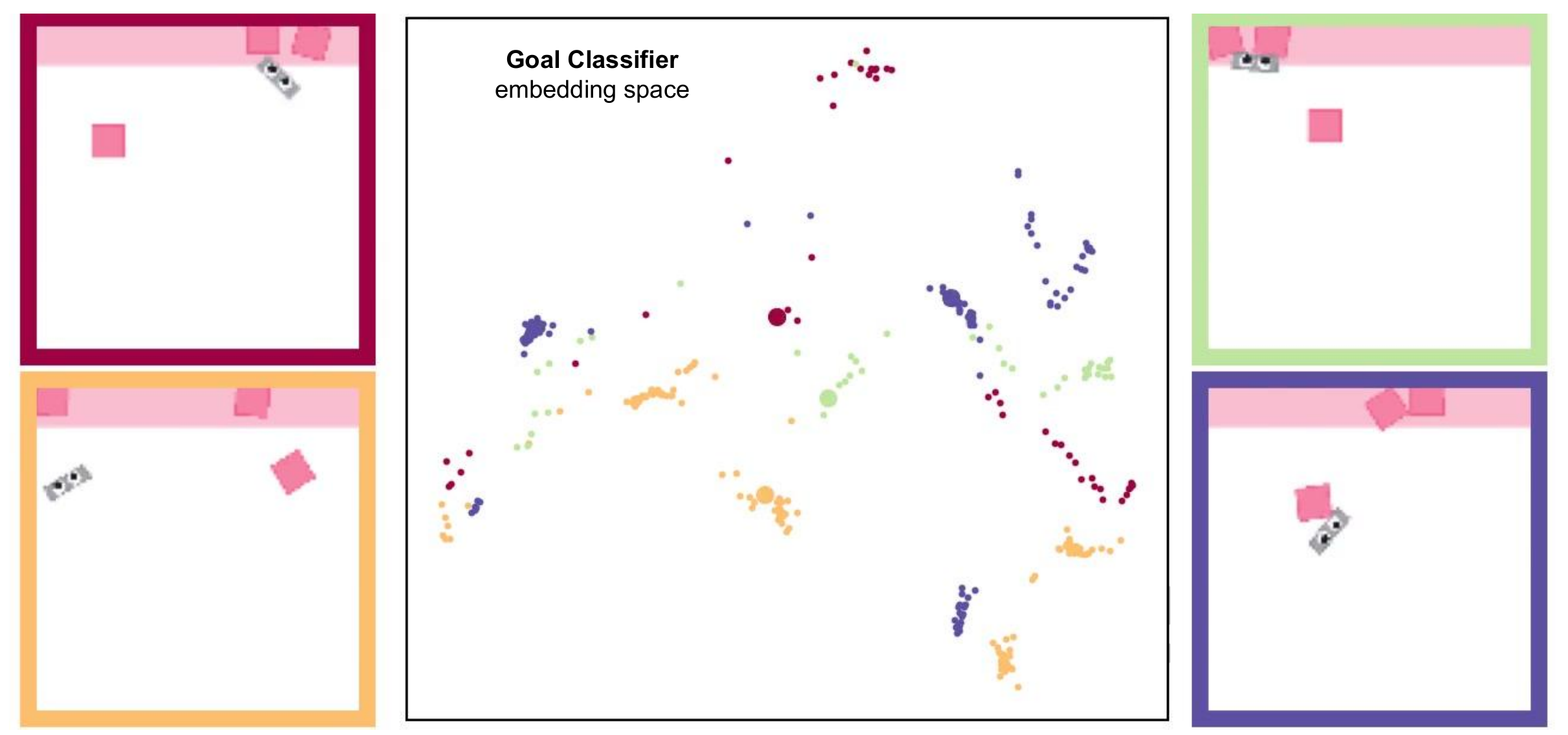}
    \caption{\textbf{t-SNE 2D projection} of the learned embedding space for XIRL (\textbf{top}) and Goal Classifier (\textbf{bottom}), where 4 video demonstrations are embedded and displayed for the \textit{shortstick} agent from the X-MAGICAL benchmark.}
    \label{fig:tsne}
\end{figure*}

To better understand and compare our learned representations, we visualize the t-SNE projection of the learned XIRL and Goal Classifier embedding spaces for 4 video demonstrations of the \textit{shortstick} agent in Figure~\ref{fig:tsne}. We observe that:
\begin{itemize}
\item Trajectories for different demonstrations overlap and are well-aligned in the XIRL embedding space. In contrast, there is significantly less structure in the Goal Classifier space.
\item Distances to the goal (top left corner in the top figure) correlate well to task progress.
\end{itemize}

We also provide video visualizations of the t-SNE embeddings in the supplementary video, which highlight some more properties.

\subsection{Nearest-Neighbor Retrieval}

We provide nearest-neighbor alignment videos in the supplementary video.

\subsection{Other Visualizations}

Please see the supplemental video for videos showing trained policy rollouts on the \textit{Sweeping} and \textit{Puck Pushing} tasks for the cross-embodiment setting, as well as interactive visualizations of the learned reward for the above environments and X-REAL.

\section{Hyperparameters}
\label{appendix:hyperparams}
In this section, we give a comprehensive overview of the hyperparameters used for representation learning and policy learning on the \textit{Sweeping} task from X-MAGICAL, the \textit{Puck Pushing} task from \cite{schmeckpeper2020reinforcement}, and X-REAL.

\subsection{Representation Learning}
\label{app:hparams_repr}

We used mostly the same hyperparameters to train the XIRL encoder across all environments. The main parameters that vary are the embedding dimension and the number of sampled frames.

\begin{table}[htbp]
    \centering
    \begin{tabular}{lcc}
        \toprule
        \textbf{Hyperparameter} & \textbf{Value} & \textbf{Range Considered}\\\midrule
        Loss & &\\
        \quad{}Type & regression (mse) & \{mse, huber, x-ent\}\\
        \quad{}Stochastic matching & False & -\\
        \quad{}Normalize time indices & True & -\\
        \quad{}Variance-aware & False & -\\
        \quad{}Distance metric & L2 & \{cosine, L2\}\\
        Optimizer & & \\
        \quad{}Type & ADAM & -\\
        \quad{}Initial learning rate & $10^{-6}$ & $10^{-5}$ to $10^{-3}$\\
        \quad{}Final learning rate & $0$& -\\
        \quad{}$\beta_1$ & $0.99$ & -\\
        \quad{}$\beta_2$ & $0.999$ & -\\
        \quad{}Weight decay & $10^{-5}$ & $10^{-6}$ to $5 \cdot 10^{-4}$\\
        Misc. & & \\
        \quad{}Total train iters. & $8\cdot10^{3}$ & $2\cdot10^{3}$ to $30\cdot10^{3}$\\
        \quad{}Batch size & $4$ & $4$ to $8$\\
        \quad{}Normalize embeddings & False & \{True, False\}\\
        \quad{}Temperature & $0.1$ & -\\
        \quad{}Learnable temperature & False & \{True, False\}\\
        \quad{}Augmentations & color jit, rand. crop, togray, gauss. blur & -\\
        \quad{}Frame Sampler & uniform & \\
        \quad{}Number of frames & & \\
        \quad{}\quad{}X-MAGICAL & $40$ & 5 to 50\\
        \quad{}\quad{}Puck Pushing & $40$ & 5 to 50\\
        \quad{}\quad{}X-REAL & $30$ & 5 to 50\\
        \quad{}Embedding size &  & \\
        \quad{}\quad{}Sweeping & $32$ & $32$ to $128$\\
        \quad{}\quad{}Puck Pushing & $64$ & $32$ to $128$\\
        \quad{}\quad{}X-REAL & $64$ & $32$ to $128$\\
        \bottomrule
    \end{tabular}
    \vspace{2mm}
    \caption{Hyperparameters for all XIRL representation learning experiments.}
    \label{tab:hp-xirl-replearn}
\end{table}

\subsection{Policy Learning}
\label{app:hparams_pl}

Most hyperparameters used for downstream reinforcement learning are identical across \textit{X-MAGICAL} and \textit{Puck Pushing}. What changes is the total number of training steps for each embodiment, since some converge much faster than others.

\begin{table}[htbp]
    \centering
    \begin{tabular}{lc}
        \toprule
        \textbf{Hyperparameter} & \textbf{Value}\\\midrule
        Total train steps &\\
        \quad{}Gripper & $500$K \\
        \quad{}Shortstick & $500$K \\
        \quad{}Mediumstick & $225$K \\
        \quad{}Longstick & $75$K \\
        \quad{}Sawyer arm (RLV env) & $200$K \\
        Optimizer & \\
        \quad{}Type & ADAM \\
        \quad{}Learning rate & $10^{-4}$\\
        \quad{}$\beta_1$ & $0.9$ \\
        \quad{}$\beta_2$ & $0.999$ \\
        Q-network & \\
        \quad{}Hidden units & $1024$ \\
        \quad{}Hidden layers & $2$ \\
        \quad{}Non-linearity &  \texttt{ReLU}\\
        Actor & \\
        \quad{}Hidden units & $1024$ \\
        \quad{}Hidden layers & $2$ \\
        \quad{}Non-linearity &  \texttt{ReLU}\\
        Misc. & \\
        \quad{}Frames stacked & $3$ \\
        \quad{}Action repetitions & $1$ \\
        \quad{}Discount factor &  $0.99$ \\
        \quad{}Minibatch size  & $1024$ \\
        \quad{}Replay period every & $1$ step \\
        \quad{}Eval period every & $5000$ step \\
        \quad{}Number of eval episodes & $50$ \\
        \quad{}Replay buffer capacity & $10^{6}$ \\
        \quad{}Seed steps & $5000$ \\
        \quad{}Critic target update frequency & $2$ \\
        \quad{}Actor update frequency & $2$ \\
        \quad{}Critic target EMA momentum ($\tau_Q$) & $0.005$\\
        \quad{}Actor log std dev. bounds & $[-5, 2]$\\
        \quad{}Entropy temperature & $0.1$\\
        \quad{}Learnable temperature & True\\
        \bottomrule
    \end{tabular}
    \vspace{2mm}
    \caption{Hyperparameters for all RL experiments.}
    \label{tab:hp-rl}
\end{table}

\end{appendices}


\end{document}